\def\year{2019}\relax
\documentclass[letterpaper]{article} 
\usepackage{aaai19}  
\usepackage{times}  
\usepackage{amssymb}
\usepackage{amsmath}
\usepackage{algorithm}
\usepackage{algorithmicx}
\usepackage{algpseudocode}
\usepackage{amsfonts}
\usepackage{helvet}  
\usepackage{courier}  
\usepackage{url}  
\usepackage{graphicx}  
\usepackage{multirow}
\usepackage{adjustbox}
\usepackage{tabu}
\usepackage{booktabs}
\usepackage{comment}
\usepackage{xcolor}
\usepackage{graphicx}
\usepackage{subfigure}
\usepackage{caption}

\begin{document}


\title{Similarity Learning with Higher-Order Graph Convolutions for \\Brain Network Analysis}



\newcommand\authorEmail[1]{#1}


\author{\fontsize{9.5}{11.4}\selectfont{Guixiang Ma\textsuperscript{1}, Nesreen K. Ahmed\textsuperscript{2}, Ted Willke\textsuperscript{2}, Dipanjan Sengupta\textsuperscript{2}, Michael W. Cole\textsuperscript{3}, Nicholas B. Turk-Browne\textsuperscript{4}, Philip S. Yu\textsuperscript{1}} \\
\textsuperscript{1}University of Illinois at Chicago, \textsuperscript{2}Intel Labs, \textsuperscript{3}Rutgers University, \textsuperscript{4}Yale University 
\\
\authorEmail{gma4@uic.edu}, \authorEmail{nesreen.k.ahmed@intel.com},
\authorEmail{ted.willke@intel.com},
\authorEmail{dipanjan.sengupta@intel.com}
\\
\authorEmail{michael.cole@rutgers.edu},
\authorEmail{nicholas.turk-browne@yale.edu},
\authorEmail{psyu@uic.edu}
}

\maketitle

\begin{abstract}
Learning a similarity metric has gained much attention recently, where the goal is to learn a function that maps input patterns to a target space while preserving the semantic distance in the input space. While most related work focused on images, we focus instead on learning a similarity metric for neuroimages, such as fMRI and DTI images. 
We propose an end-to-end similarity learning framework called \emph{Higher-order Siamese GCN} for multi-subject fMRI data analysis. The proposed framework learns the brain network representations via a supervised metric-based approach with siamese neural networks using two graph convolutional networks as the twin networks. Our proposed framework performs higher-order convolutions by incorporating higher-order proximity in graph convolutional networks to characterize and learn the community structure in brain connectivity networks. To the best of our knowledge, this is the first community-preserving similarity learning framework for multi-subject brain network analysis. Experimental results on four real fMRI datasets demonstrate the potential use cases of the proposed framework for multi-subject brain analysis in health and neuropsychiatric disorders. Our proposed approach achieves an average AUC gain of $75$\% compared to PCA, an average AUC gain of $65.5$\% compared to Spectral Embedding, and an average AUC gain of $24.3$\% compared to S-GCN across the four datasets, indicating promising application in clinical investigation and brain disease diagnosis.
\end{abstract}

\section{Introduction}
\label{sec:intro}

Learning a similarity metric has gained much attention recently in a variety of real-world applications, where the goal is to learn a function that maps input patterns to a target space while preserving the semantic distance in the input space~\cite{wang2014learning,chopra2005learning,guo2001learning,chechik2009online,bautista2017deep}. For example, finding images that are similar to a query image is an indispensable problem in search engines \cite{wang2014learning}, and an effective image similarity metric is the key for finding similar images. Similarity learning is particularly suitable for applications where the number of classes is very large, the number of samples is very small, and/or only a subset of classes is known at the time of training (e.g., face recognition applications). Metric learning systems are typically trained on pairs of patterns from the training set (e.g., pairs of images, graphs, etc), with the goal to find a function that maps the input patterns to a target space such that the distance among patterns in the target space approximates the semantic distance in the input space. Given a family of possible models (functions) with parameter weights $\mathbb{W}$, the goal is to find the value of optimal weights $\mathbb{W}$, such that the estimated similarity score of the input pair $(X_1, X_2)$ is small if $X_1$ and $X_2$ belong to the same class, and large if they belong to different classes. The loss function used in training maximizes the mean similarity between input pairs belonging to the same class, and minimizes the mean similarity between pairs belonging to different classes.   


While  the majority of the similarity learning research is focused on images, our work focused instead on learning a similarity metric for neuroimages, such as fMRI images and DTI images. Functional magnetic resonance imaging (fMRI) measures brain activity by detecting changes associated with blood flow, and has been widely used in cognitive neuroscience, medical, and clinical applications. We particularly focus on brain connectivity networks (i.e., functional networks) extracted from the input fMRI images, where the observed brain activity is modeled as a network of inter-regional functional associations. The resulting functional networks can be analyzed/modeled using graph-theoretic and graph machine learning techniques. The analysis of brain connectivity networks has revealed many key features of the brain organization, such as the modular structure of the brain, and the presence of hubs \cite{bertolero2015modular,power2011functional}.   


Graph convolutional neural networks (GCNs) have emerged to learn deep representations of graph-structured data, 
and been shown to outperform other traditional relational learning methods~\cite{defferrard2016convolutional,li2018deeper}. However, these works focus mainly on social and information networks, to model the node/graph similarity, where the relationships/links are usually given or well-defined (e.g., friendship, retweet, following links). In this paper, however, we focus on learning deep representations from fMRI brain connectivity networks, where each brain network represents the brain activity patterns of a particular subject. Brain connectivity networks are widely used to model inter-regional functional connections, and are typically inferred from the input fMRI data of each subject. Each brain connectivity network is usually represented by a graph constructed from input fMRI data where the nodes represent brain regions and the edges represent the pairwise region-to-region correlations. However, the organization of brain connectivity networks is constrained by various factors, such as the underlying brain anatomical network~\cite{hagmann2008mapping,honey2009predicting}, which has an important contribution in shaping the activity across the brain. These constraints pose significant challenges on characterizing the structure and organization of brain connectivity networks. In addition, these constraints lead to distance-dependent inter-regional correlations, where activity patterns from nearby brain regions is strongly correlated due to spatial factors than the activity from distant ones. Thus, it's crucial to distinguish between long-range functional connectivity that represents the brain inter-regional communications, which arise from functional necessity, from those that arise due to spatial constraints~\cite{honey2009predicting}. We propose an end-to-end similarity learning framework for multi-subject fMRI data. The proposed framework learns the brain network representations via a supervised metric-based approach with siamese neural networks and using two GCNs as the twin networks. Our proposed framework performs higher-order convolutions by incorporating higher-order proximity via random walks in graph convolutional networks. We use this approach to characterize the community and organizational structure in brain connectivity networks, and to emphasize the contribution of the long-range functional connectivity. To the best of our knowledge, this is the first community-preserving similarity learning framework for multi-subject brain network analysis. 

The principle aim of this work is to build a framework that can learn discriminative structural features of brain connectivity networks for multi-subject analysis. However, the complex structure of these networks poses significant challenges to such a framework. For example, prior work has shown the importance of higher-order proximity structure such as the community structure of brain networks in health as well as neuropsychiatric disorders~\cite{ma2016multi,ma2017multi}. However, current graph neural methods are capable of only learning local neighborhood information and cannot directly capture the higher-order proximity and community structure of graphs. Although it might be possible for GCNs to learn these higher-order patterns under certain conditions, for example, stacking more hidden GCN layers to allow more information to flow across the graph, these conditions are unlikely to work well in practice, which is due to the non-linearity of the GCN layers, and the difficulty of fitting the larger number of parameters with a limited number of training samples (e.g., the limited samples in neuroimaging data) which leads to overfitting (see Figures~\ref{fig:acc_K} and~\ref{fig:acc_Layers}). Thus, how to generalize GCNs to capture the community structure of brain networks is a key challenge. Moreover, how to leverage the structural features learned by GCNs for similarity learning and build an end-to-end similarity learning framework for multi-subject analysis is another challenging problem. Our proposed framework leverages higher-order graph convolutional networks to characterize the community and organizational structure of brain connectivity networks. We conjecture this framework would be useful for the multi-subject brain analysis in health and neuropsychiatric disorders~\cite{greicius2008resting}. Our main contributions are summarized as follows:
\begin{itemize}
\setlength{\itemsep}{0pt}
    \item We propose an end-to-end similarity learning framework called "Higher-order Siamese GCN" for metric learning of graph data. The framework learns the brain network representations via a supervised metric-based approach with siamese neural networks and using two GCNs as the twin networks. The framework leverages higher-order graph convolutions to characterize the community structure of brain connectivity networks.

    \item We apply the proposed framework on four real fMRI brain network datasets for similarity learning with respect to brain health status and cognitive abilities. The experiment results demonstrate the effectiveness and advantage of the proposed framework for neurological disorder analysis with respect to three kinds of disease (i.e., Autism, Bipolar and HIV) and the cognitive analysis for the Human Connectome Project. 
    
    \item Our proposed approach achieves an average AUC gain of $75$\% compared to PCA, $65.5$\% compared to Spectral Embedding, and $24.3$\% compared to S-GCN across a variety of datasets, indicating its promising use cases for multi-subject brain analysis in health and neuropsychiatric disorders.
\end{itemize}
\vspace{-2mm}


\section{Similarity Learning Framework for Multi-Subject fMRI Data}
\label{sec:framework}
\begin{figure*}[t]
\centering
    \begin{minipage}[l]{0.7\linewidth}
      \centering
      \includegraphics[width=0.9\linewidth]{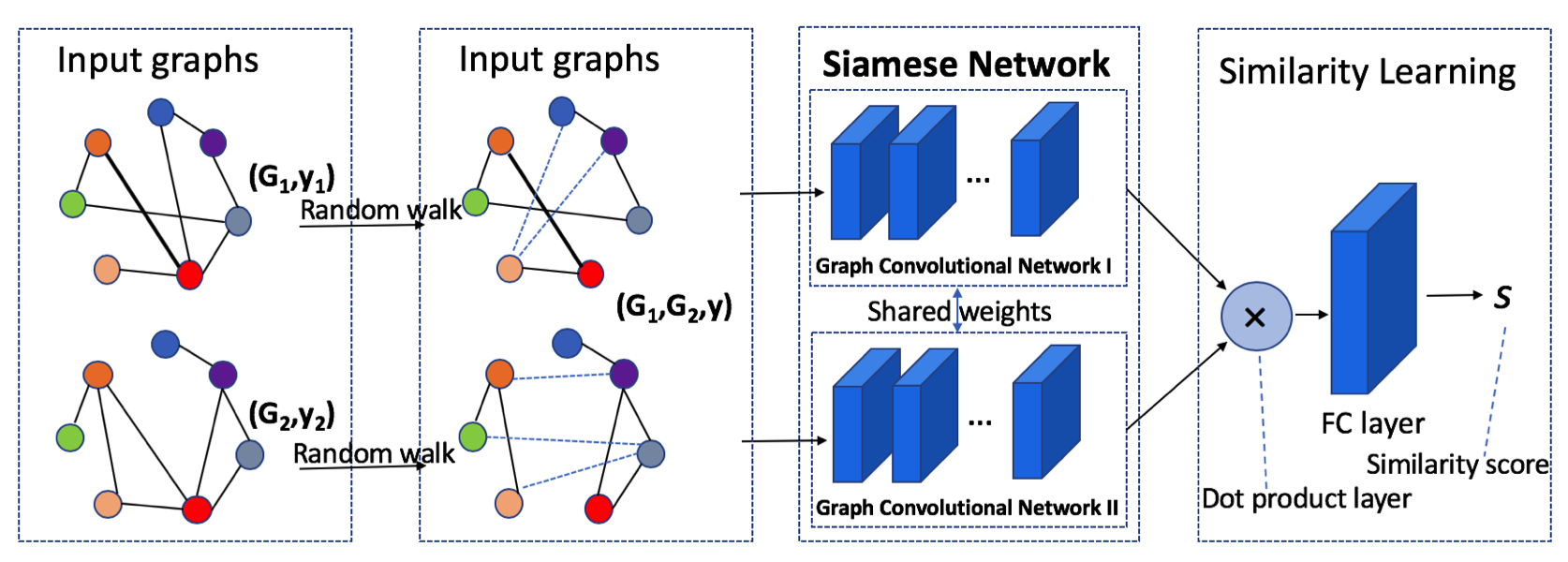}
    \end{minipage}
  \caption{The Higher-order Siamese-GCN framework}\label{fig:framework}
  \vspace{-4mm}
\end{figure*}

We propose an end-to-end framework called \emph{higher-order Siamese GCN} for learning the similarity among fMRI brain connectivity networks extracted from multiple subjects. We assume each brain network is represented by a graph where the nodes represent brain regions and the edges represent the inter-regional associations. Our proposed framework (presented in Figure~\ref{fig:framework}) consists of the following: (1) a Siamese neural network architecture, (2) two identical higher-order graph convolutional networks used as the twin network, (3) a dot product layer, and (4) a fully connected layer. 
The framework takes an input pair of graphs, such that each graph is processed by one branch of the siamese network (i.e., each graph is processed by one of the identical twin GCNs). Since the twin graph neural networks share the same weights, each input graphs is guaranteed to be processed in the same way by the twin networks respectively. As such, similar input graphs would be embedded similarly in the latent space. Our proposed framework performs higher-order convolutions by incorporating higher-order proximity using random walks in graph convolutional networks. Finally, the dot product layer combines the outputs from the two branches of the siamese network, then followed by a single fully connected (FC) layer with one output to estimate the similarity score.

Given a multi-subject fMRI data set $\mathcal{G} = \{G_1,G_2,\cdots ,G_N\}$ where $G_i=(V_i, E_i, \mathbf{A_i})$ is the fMRI brain network of subject $i$, $V_i$ is the set of vertices in $G_i$, $E_i \subset V_i \times V_i $ is the set of edges in $G_i$, and $\mathbf{A_i} \in \mathbb{R}^{m \times m}$ is the affinity matrix of $G_i$, we assume that we have a model $\mathcal{M}$ with parameter weights $\mathbb{W}$ to map the brain networks to the feature/embedding space and the embedding of $G_i$ can be obtained as $H_i = \mathcal{M}(A_i)$. Assume we have a function $\mathcal{S}$ for computing similarity between the embeddings of two brain networks, and the similarity score between a pair of brain networks $G_i$ and $G_j$ is denoted by $s_{ij} = \mathcal{S}(H_i,H_j)$, the goal of similarity learning for the multi-subject fMRI data is to learn the optimal parameter weights $\mathbb{W}$ for $\mathcal{M}$ such that the value of $s_{ij}$ is large for an input pairs $(G_i, G_j)$ belonging to the same class, and small for input pairs $(G_i, G_j)$ belonging to different classes. The loss function used in training should maximize the mean similarity between input pairs belonging to the same class, and minimize the mean similarity between pairs belonging to different classes.


\vspace{-1mm}
\paragraph{\textbf{From fMRI data to brain connectivity networks.}}
A whole-brain fMRI image consists of a sequence of 3D brain image scans, where each volume consists of hundreds of thousands of voxels. Each voxel has an intensity value that is proportional to the strength of the Nuclear Magnetic Resonance (NMR) signal emitted at the corresponding location in the brain volume. The  raw form of an fMRI brain image is a discrete time series of 3D images. In order to convert the original fMRI images to region-by-region brain networks, we extract a sequence of responses from each of the regions of interest (ROI), where each ROI represents a brain region. We perform the standard fMRI brain image processing steps, including functional images realignment, slice timing correction and normalization, etc. Then, we compute the region-to-region brain activity correlations. We only keep the positive correlations as the links among the brain regions. The final constructed network is a graph where the nodes/vertices represent brain regions and the edges are the region-to-region correlations.

\vspace{-1mm}
\paragraph{\textbf{Graph Convolutional Networks.}}
Graph convolutional network (GCN), as a generalized convolutional neural network from grid-structure domain to graph-structure domain, has been emerging as a powerful approach for graph mining \cite{bruna2013spectral,defferrard2016convolutional}. In GCNs, filters are defined in the graph spectral domain. In this paper, we employ the GCN with fast spectral filtering proposed in \cite{defferrard2016convolutional}. Given a graph $G=(V, E, \mathbf{A})$, where $V$ is the set of vertices, $E \subset V \times V $ is the set of edges, and  $\mathbf{A} \in \mathbb{R}^{m \times m}$ is the adjacency matrix, the diagonal degree matrix $\mathbf{D}$ will have elements $\mathbf{D}_{ii} = \sum_j \mathbf{A}_{ij}$. The graph Laplacian matrix is $\mathbf{L} = \mathbf{D} - \mathbf{A}$, which can be normalized as $\mathbf{L} = \mathbf{I}_m - \mathbf{D}^{-\frac{1}{2}}\mathbf{A} \mathbf{D}^{-\frac{1}{2}}$, where $\mathbf{I}_m$ is the identity matrix. Assume the orthonormal eigenvectors of $\mathbf{L}$ are represented as $\{u_l\}_{l=0}^{m-1}\in \mathbb{R}^{m \times m}$, and their associated eigenvalues are $\{\lambda_l\}_{l=0}^{m-1}$, the Laplacian is diagonalized by the Fourier basis $[u_0, \cdots,u_{m-1}](=\mathbf{U})\in \mathbb{R}^{m \times m}$ and $\mathbf{L} = \mathbf{U\Lambda U^T}$ where $\mathbf{\Lambda} = diag([\lambda_0,\cdots,\lambda_{m-1}])\in \mathbb{R}^{m\times m}$. The graph Fourier transform of a signal $x\in \mathbb{R}^m$ can then be defined as $\hat{x} = \mathbf{U^T}x \in \mathbb{R}^m$\cite{shuman2013emerging}. Suppose a signal vector $\mathbf{x} : V \rightarrow \mathbb{R}$ is defined on the nodes of graph $G$, where $\mathbf{x}_i$ is the value of $\mathbf{x}$ at the $i^{th}$ node. Then the signal $\mathbf{x}$ can be filtered by $g_\theta$ as

\begin{align}
    y = g_\theta*\mathbf{x} = g_\theta(\mathbf{L})\mathbf{x} = g_\theta(\mathbf{U{\Lambda}U^T})\mathbf{x} = \mathbf{U}g_\theta(\Lambda)\mathbf{U^T}\mathbf{x}
\label{eq:filter_signal}
\end{align}
where the filter $g_\theta(\Lambda)$ can be defined as $g_{\theta}(\Lambda) = \sum_{k=0}^{K-1}{\theta_k}{\Lambda^k}$, and the parameter $\theta\in {\mathbb{R}}^K$ is a vector of polynomial coefficients \cite{defferrard2016convolutional}. The above filter is exactly $K$-localized, which means the nodes with shortest path length greater than $K$ are not considered for the convolution. To further reduce computational complexity,  we use the Chebyshev polynomials \cite{defferrard2016convolutional} which can be computed recursively by $T_k(\mathbf{x}) = 2\mathbf{x}T_{k-1}(\mathbf{x}) - T_{k-2}(\mathbf{x})$ with $T_0 = 1$ and $T_1 = \mathbf{x}$, and a filter of order $K-1$ is parameterized as the truncated expansion 
\vspace{-2mm}
\begin{align}
    g_{\theta}(\Lambda) = \sum_{k=0}^{K-1}{\theta_k}T_k(\hat{\Lambda})
\end{align}
Then the filtering operation can be written as $y = g_\theta(\mathbf{L})\mathbf{x} = \sum_{k=0}^{K-1}\theta_kT_k(\hat{L})\mathbf{x}$, where $T_k(\hat{L})\in \mathbb{R}^{n\times n}$ is the Chebyshev polynomial of order $k$ with the Laplacian $\hat{L} = 2L/\lambda_{max} - I_n$. The $j^{th}$ output feature map of sample s is then given by
\vspace{-2mm}
\begin{align}
    y_{s,j} = \sum_{i=1}^{F_{in}}g\theta_{i,j}(L)x_{s,i}\in \mathbb{R}^m
\label{eq:filter}
\end{align}
where $x_{s,i}$ is the input feature map, and $F_{in}$ represents the number of input filters. The $F_{in}\times F_{out}$ vectors of Chebyshev coefficients $\theta_{i,j}\in \mathbb{R}^K$ are the layer's trainable parameters. Then we build the GCN by stacking multiple convolutional layers in the form of Equation (\ref{eq:filter}), with a non-linearity activation (ReLU) following each layer. 

\paragraph{\textbf{Higher-order Graph Convolutional Networks.}}
As discussed above, GCN networks use spectral filterings, which consider localized convolutions while ignoring vertices with shortest path length beyond a threshold. This limits the ability of traditional GCNs to learn the complex structural patterns of graphs, such as the community structure, motif patterns, higher-order proximity, etc. Recently, high-order structural patterns were shown to be crucial for learning more accurate graph representations~\cite{ahmed2015efficient,rossi2018higher,benson2016higher,rossi2017deep}. 

Thus, we modify the traditional GCN model to operate directly on graphs with higher-order proximity representations so that the higher-order information will be incorporated during the convolutions. Specifically, we employ a random walk sampling process on graphs to obtain the higher-order proximity representations. Random walk sampling has been widely used for learning graph embedding that capture the higher-order proximity and community structure in the graph~\cite{perozzi2014deepwalk,ahmed2018learning}. In our framework, the random walk strategy helps incorporate the higher-order structural information into the graph representations, which further allows for the higher-order graph convolutions in GCNs to capture community and organizational structure of brain networks.

We denote a random walk rooted at vertex $v_i$ as $W_{v_i}$, which is a stochastic process with random variables $W_{v_i}^1, W_{v_i}^2, \cdots, W_{v_i}^n$, and $W_{v_i}^{j+1}$ is a vertex chosen randomly from the neighbors of vertex $W_{v_j}$. Given a graph $G$, the random walk generator samples uniformly a random vertex $v_i$ as the root of the random walk. The walk uniformly samples a vertex from the neighbors of the root, after which it continues sampling from the neighbors of the last vertex visited until the maximum walk length is reached. There could be multiple walks starting from each vertex, depending on the number of walks specified\cite{perozzi2014deepwalk}. Line $2-9$ in Algorithm \ref{alg:Multihop_SGCN} illustrates the random walk sampling we use for capturing higher-order proximity information. Note that we slide a window with size $w$ on each walk generated and record the frequency of nodes that co-occur within a window in $\mathbf{F}$, and $w$ is a parameter that controls the extent of the higher-order proximity considered in the graph representation. After we obtain the adjacency matrix $\mathbf{A}$ in Line $12$ of Algorithm \ref{alg:Multihop_SGCN} for the higher-order proximity representation, we use $\mathbf{A}$ for computing the Laplacian matrix $\mathbf{L}$, and use this $\mathbf{L}$ for the spectral filtering in Equation (\ref{eq:filter_signal}) to enable the higher-order graph convolutions.
\vspace{-1mm}
\paragraph{\textbf{Algorithm Description.}}
Algorithm \ref{alg:Multihop_SGCN} shows the overall process of the Higher-order Siamese GCN approach. As the spectral graph convolutional networks filter signals are defined on a common graph Laplacian for all samples, we first estimate the mean functional connectivity matrix among the training samples by computing the mean affinity matrix over the graphs and obtain the k-nn graph $\Bar{G}$ using the correlation distance between region pairs. Then we obtain the binary adjacency matrix $\Bar{\mathbf{A}}$ for $\Bar{G}$ as its graph representation. In order to obtain the higher-order proximity representation of the graph, we apply random walk on $\Bar{G}$, after which we slide a window with size $w$ on each walk to get the co-occurrence frequency between two nodes and record the frequency in the matrix $F$. After the random walk sampling stage, we obtain the $k$-nn graph $G'$ based on $F$, where the edges encode the higher-order proximity of the original graph $\Bar{G}$. Then we merge the edges of $G'$ into $\Bar{G}$ to get the updated graph representation with higher-order proximity and obtain the adjacency matrix of the updated graph as $\mathbf{A}$. We compute the Laplacian matrix of the new graph representation and denote it with $\mathbf{L}$. Now we start the model learning of Higher-order Siamese GCN. We first prepare the pairs of training samples from $\mathcal{G}$ with label $Y_{ij} = 1$ for each pair of same class and $Y_{ij} = -1$ for each pair with different classes. We also initialize the neural network parameters $\Theta$ of the GCNs with Siamese network. Then we input the pairs into the Siamese GCNs and perform spectral convolutions according to Equation (\ref{eq:filter}). Specifically, for each input brain network, we use the functional correlation value of node $v_s$ with $v_i$ as the signal $x_{s,i}$ in Equation (\ref{eq:filter}). After the convolutions, the outputs of the twin GCNs are then combined by a dot product layer, followed by a fully connected (FC) layer that integrates the information learned from the preceding filters and output a similarity estimiate $s_{ij}$ for the input pair $G_i$ and $G_j$. Then we compute the loss for the Siamese network. We use the Hinge loss in Equation (\ref{hinge_loss}), by minimizing which the mean similarity between input pairs belonging to the same class will be maximized while the mean similarity between pairs belonging to different classes will be minimized.   We use stochastic gradient descent with Adaptive Moment Estimation (ADAM) optimizer to train the model.
\vspace{-0.3em}
\begin{align}
    L^{hinge} = \frac{1}{N_p}\sum_{i=1}^{N}\sum_{j=i+1}^{N} max(0,1-{Y_{ij}}{s_{ij}}),
    \label{hinge_loss}
\end{align}
where $N$ is the total number of subjects in the training set, and $N_p = N(N-1)/2$ is the total number of pairs from the training set.

\renewcommand{\algorithmicrequire}{\textbf{Input:}}
\renewcommand{\algorithmicensure}{\textbf{Output:}}
\begin{algorithm}[t]
\caption{Higher-order Siamese GCN}
\label{alg:Multihop_SGCN}
\begin{algorithmic}[1] 
\scriptsize
\Require $\mathcal{G} = {G_1, G_2, \cdots, G_n}$ (training graph samples); $\mathbf{y}$(class labels); random walk parameters: $\gamma$ (number of walks), $l$ (walk length), $w$ (window size)\\
    Obtain the mean $k$-nn graph $\Bar{G}(V,E,\Bar{\mathbf{A}})$;\\
    Initialize a frequency matrix $\mathbf{F}\in \mathbb{R}^{m\times m}$ with $0$s;
    \For {$i = 0$ to $\gamma$}
    \State $V'=$ Shuffle$(V)$;
    \For{\textbf{each} $v_i \in V'$}
    \State $W_{v_i} = RandomWalk(\Bar{G},v_i,l)$;
    \State Update $\mathbf{F}$;
    \EndFor
    \EndFor\\
    Obtain a $k$-nn graph $G'$ based on $\mathbf{F}$;\\
    Merge the edges of $G'$ into $\Bar{G}$; \\
    Obtain the updated adjacency matrix $\mathbf{A}$;\\
    Prepare pairs of training samples from $\mathcal{G}$;\\
    Initialize the parameters $\Theta$ of GCNs in Siamese network;
    \While {not converge}
    \State Perform spectral filterings according to Equation (\ref{eq:filter}); 
    \State Compute the similarity estimate $s_{ij}$ for the input pair $(G_i,G_j)$;
    \State Compute the loss $L^{hinge}$ according to Equation (\ref{hinge_loss}) ;
    \State Apply stochastic gradient descent with ADAM optimizer to update $\Theta$;
    \EndWhile
\end{algorithmic}
\end{algorithm}


\vspace{-0.1em}

\section{Experiments \& Results}
We evaluate the performance of the proposed framework on four real resting-state fMRI brain datasets and we compare with the state-of-the-art baselines. There are mainly two categories of neuroimaging applications involved in the evaluation: (1) medical imaging analysis, which aims to distinguish brain disordered subjects and healthy controls; (2) the cognitive analysis using Human Connectome Project data \cite{van2013wu}, where the goal is to learn similarity with respect to cognitive abilities from the brain networks. It is worth mentioning that the empirical similarity learning on fMRI brain networks for cognitive analysis is seldom studied so far \cite{bookheimer2018lifespan} and our work provides new insights and capabilities into the problems in this area. 
\vspace{-2.mm}
\subsection{Datasets and Preprocessing}
\begin{figure}[t]
\centering
    \begin{minipage}[0.6]{1\linewidth}
      \centering
      \includegraphics[width=0.6\linewidth]{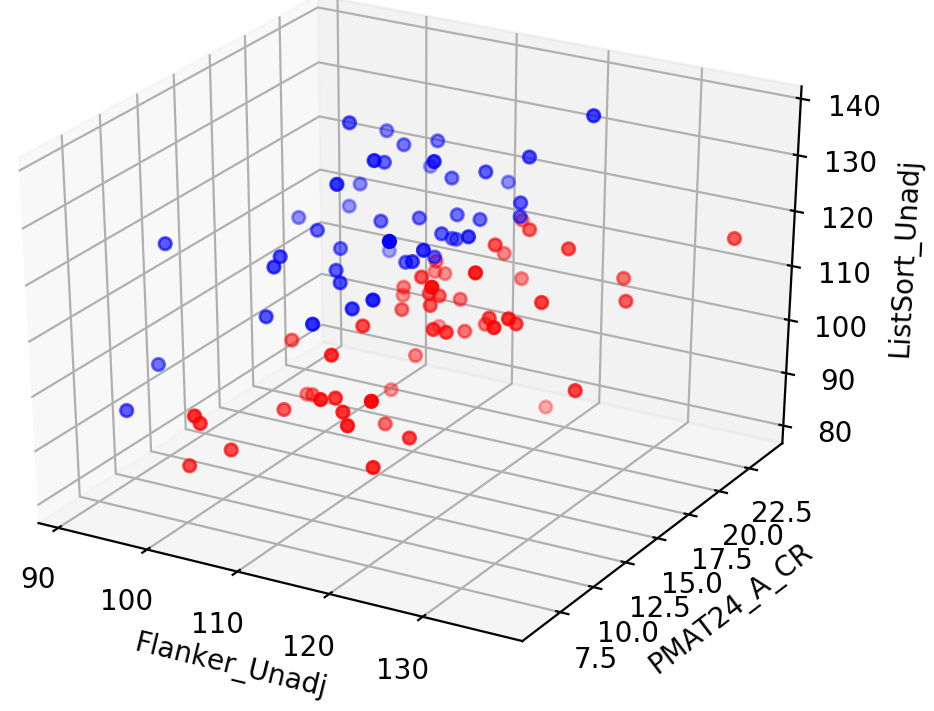}
    \end{minipage}
    \vspace{-3mm}
  \caption{The 2 clusters of HCP subjects based on three key cognitive features.}\label{fig:hcp_cluster}
\end{figure}

\begin{itemize}
    \item \emph{Autism Brain imaging Data Exchange (ABIDE)}: This dataset is provided by the ABIDE initiative \cite{di2014autism}. It has the resting-state fMRI images of 70 patients with autism spectrum disorder (ASD) and 102 healthy controls, acquired from the largest data acquisition site involved in that project. The preprocessing of the fMRI data includes slice timing correction, motion correction, band-pass filtering and registering to standard anatomical space. After the preprocessing, a brain network with 264 nodes was constructed for each subject by computing the pearson correlation between the fMRI time series of the 264 putative regions.
    \item \emph{Human Connectome Project (HCP)}: This dataset consists of resting-state fMRI imaging data and behavioral data of 100 healthy volunteers from the publicly available Washington Univeristy - Minnesota (WU-Min) Humman Connectome Project (HCP) \cite{van2013wu}. The preprocessing of the fMRI data consists of intensity normalization, phase-encoding direction unwarping, motion, correction, spatial normalization to standard template and artifact removing\cite{spronk2018mapping}. After preprocessing, for each subject, BOLD time series were extracted from the 360 parcels, and functional connectivity network with 360 nodes was constructed for each individual. In this work, we are interested in solving the pair classification problem with respect to cognitive traits. Since this dataset does not have class labels of cognitive traits, we use three key cognitive features from the participants' behavioral data, including Executive Function (Flanker Task), Fluid Intelligence (Penn Progressive Matrices) and Working Memory (List Sorting) \footnote{\url{https://www.humanconnectome.org/study/hcp-young-adult/}}, and apply K-means clustering with the three features to cluster the subjects into 2 groups. Figure~\ref{fig:hcp_cluster} shows the plot of the 2 clusters in the 3-dimension feature space, where there is a clear boundary between the two clusters. We label the subjects in the two clusters as Class 1 and Class 2, respectively, and prepare pairs of subjects based on this class label for the pair classification evaluation.    
    \item \emph{Bipolar}:  This dataset consists of the fMRI data of 52 bipolar I subjects who are in euthymia and 45 healthy controls with matched demographic characteristics \cite{cao2015identification}. The brain networks were constructed with the CONN\footnote{http://www.nitrc.org/projects/conn} toolbox \cite{whitfield2012conn}. The raw images were realigned and co-registered, followed by the normalization and smoothing steps. Then the confound effects from motion artifact, white matter, and CSF were regressed out of the signal. Finally, the brain networks were created using the signal correlations between each pair of regions among the 82 labeled Freesurfer-generated cortical/subcortical gray matter regions.
    
    \item \emph{Human Immunodeficiency Virus Infection (HIV)}: This dataset is collected from the Chicago Early HIV Infection Study at Northwestern University\cite{ragin2012structural}. It contains the resting-state fMRI data of 77 subjects, 56 of which are early HIV patients and the other 21 subjects are seronegative controls. We use the DPARSF toolbox\footnote{http://rfmri.org/DPARSF.} to process the fMRI data. The images were realigned to the first volume, followed by the slice timing correction and normalization. We focus on the 116 anatomical volumes of interest (AVOI) and construct a brain network with the 90 cerebral regions, where each node in the graph represents a cerebral region, and links are created based on the correlations between different regions.
\end{itemize}
\subsection{Baselines and Metrics}
We compare our Higher-order Siamese GCN framework with three other baseline methods for two classification tasks based on the similarity learning on brain networks, and we use AUC and accuracy as the evaluation metrics. 
\vspace{-2mm}
\begin{itemize}
    \item \textbf{PCA} is the Principal Component Analysis approach that is widely used for dimension reduction and feature extraction \cite{smith2002tutorial}. We apply PCA on the correlation matrices of the brain networks and perform similarity learning based on the PCA results. 

    \item \textbf{SE} is the Spectral Embedding approach, which finds a low dimensional representation of the data using a spectral decomposition of the graph Laplacian \cite{belkin2003laplacian}. 

    \item \textbf{Siamese GCN (S-GCN)} is the method proposed in \cite{ktena2018metric}, which employs the traditional GCN to capture localized structural information of graphs and learn similarity scores between graphs based on the outputs of GCNs in a Siamese network.This was the first work of applying graph convolutional neural network on brain connectivity networks.

    \item \textbf{Higher-order Siamese GCN (HS-GCN)} is the proposed approach in this paper. 
\end{itemize}
\vspace{-4mm}
\paragraph{\textbf{Experimental Setup.}} We use $60\%$ of the data for training and the other $40\%$ for testing. We use the experiment settings for the S-GCN following the instructions provided in \cite{ktena2018metric} and use $5$-fold cross validation for optimizing the parameters in both S-GCN and HS-GCN. For HS-GCN, we use $2$ GCN layers with $f=32$ features for each. We use the stochastic gradient descent with ADAM algorithm \cite{kingma2014adam} for the optimization, where we set learning rate to be $0.001$ and use $K = 3$ for the polynomial filters in the spectral filtering. We set the dropout rate at the fully connected layer as $0.8$ and use $0.0005$ for the regularization parameter. For the constrained variance loss in Equation (\ref{eq:convar_loss}) used in the subject classification task, we set $a = m/2$ for both datasets. For the parameters in random walk, we employ the grid search in a range of values to find the optimal parameter values. We fix the number of walks $\gamma$ to be $10$, and search the value for the walk length $l$ from $[30,40,\dots,100]$ and the value for window size $w$ from $[1,2,\dots,10]$. For the $k$-nn graph construction stage, we use $10\%$ of the number of nodes in the brain networks as the value for $k$. For PCA and SE, the only parameter is the number of components to be preserved in the output lower dimensional representation. We employ the grid search in the range of $[5,10,15,\dots,60]$ to find the optimal value for that parameter. After we obtain the output representations from PCA and SE, we calculate the similarity score for each pair according to Equation (\ref{eq:distance}) \cite{frey2007clustering}.
\vspace{-2mm}
\begin{align}
\small
    {s_{ij}} = 1 - \sqrt{\text{Tr} \left( \left(\mathbf{F}_i - \mathbf{F}_j \right)^{\mathrm{T}} \left(\mathbf{F}_i - \mathbf{F}_j \right) \right)}
    \label{eq:distance}
\end{align}
where $\mathbf{F}_i$ and $\mathbf{F}_j$ are the PCA results of subject $i$ and $j$, respectively. For each experiment, we run for $5$ times and report the average results.

\subsection{Evaluations}
We evaluate the performance of the proposed framework in similarity learning of brain networks by applying it in two classification tasks: (1) Pair classification, and (2) Subject classification.

\begin{table}[h]
\caption{AUC Scores of Pair Classification (mean $\pm$ std).}
\label{tab:results}
\centering
\resizebox{0.8\columnwidth}{!}{
\begin{tabular}{lcccc}
\toprule
Methods        &\emph{ABIDE}   &\emph{HCP}  &\emph{HIV}  &\emph{Bipolar}\\
\midrule
PCA
&$0.51 \pm 0.01$      &$0.52 \pm 0.01$  &$0.54 \pm 0.07$  &$0.52 \pm 0.01$\\
SE
&$0.55 \pm 0.02$      &$0.54 \pm 0.01$  &$0.57 \pm 0.02$  &$0.55 \pm 0.01$\\
S-GCN
&$0.78 \pm 0.29$      &$0.81 \pm 0.36$  &$0.61 \pm 0.25$  &$0.74 \pm 0.19$\\
HS-GCN
&\textbf{0.96} $\pm$ \textbf{0.02}  &\textbf{0.98} $\pm$ \textbf{0.03}  &\textbf{0.77} $\pm$ \textbf{0.20}  &\textbf{0.94} $\pm$ \textbf{0.07}\\
\bottomrule
\end{tabular}}
\end{table}

\begin{figure}[h]
\centering
    \begin{minipage}[0.6]{1\linewidth}
      \centering
      \includegraphics[width=0.7\linewidth]{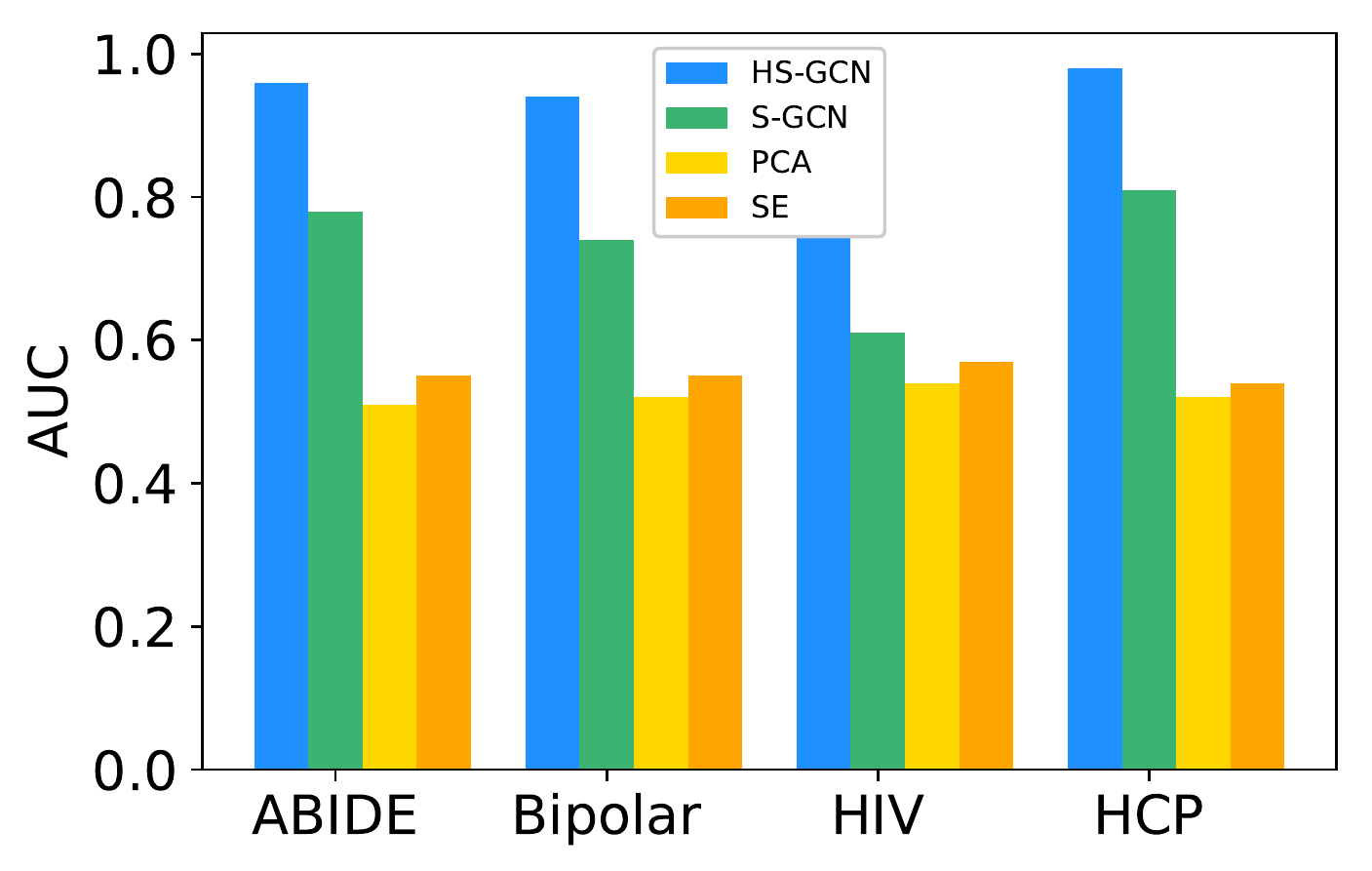}
    \end{minipage}
    \vspace{-3mm}
  \caption{AUC Scores of Pair Classification.}\label{fig:auc_bar}
\end{figure}

\paragraph{\textbf{Pair Classification.}}
Pair classification refers to the classification of similar pairs (brain networks from the same class) versus dissimilar pairs (brain networks from different classes) based on the similarity learned by the model. This is a very important task in brain connectivity analysis, especially for the brain disorder identification problem when there is very limited number of labeled samples, which is a common scenario in biomedical data mining. If we could build a powerful model that can distinguish similar pairs and dissimilar pairs well, then we could use it for predicting unseen samples by looking into its similarity with the labeled samples. With the Siamese architecture, the proposed model has inherent advantages for this task. For instance, given a training set with $100$ samples, we could generate $4950$ unique pairs of brain networks that could be used as inputs for the Higher-order Siamese GCN model, which would greatly guarantee the training effectiveness and robustness of the model when applying it for predicting unseen samples. Table \ref{tab:results} and Figure~\ref{fig:auc_bar} shows the classification AUC of the Higher-order Siamese GCN and that of the baseline methods on pair classification on the four datasets. 

As shown in Table \ref{tab:results}, the classification AUC of HS-GCN is significantly higher than that of the baseline methods. More specifically, our proposed HS-GCN achieves an average AUC gain of $75$\% compared to PCA, an average AUC gain of $65.5$\% compared to Spectral Embedding, and an average AUC gain of $24.3$\% compared to S-GCN across all datasets. Our proposed HS-GCN is more accurate and has a lower variance compared to S-GCN. 
The PCA based approach achieved the lowest AUC scores. This is probably due to the fact that PCA learns lower dimensional feature representations directly from the correlation matrix while not considering the structural information of the graph. 
The Spectral Embedding based approach achieved higher AUC than PCA. Since SE is designed for spectral clustering which means the embedding results of SE encodes the community structure of graphs, this property would help discriminate the brain networks of different classes. But since SE is not able to capture the complex local structure of brain networks and it is not an end-to-end approach for similarity learning, the AUC scores of SE is much lower than the proposed approach. The S-GCN model achieved fairly good results on ABIDE and HCP datasets but the AUC scores are still not as good as that of HS-GCN. The superior performance of our proposed HS-GCN framework indicates that the localized structural information and community structure learned by higher-order graph convolutions did benefit the similarity learning of brain networks. Moreover, the HS-GCN achieved the best results on all the four datasets, demonstrating its generalizing ability in similarity learning of brain networks. 

\begin{figure}[h]
\centering
        \begin{subfigure}[\scriptsize {Bipolar}]{		\includegraphics[width=.45\linewidth] {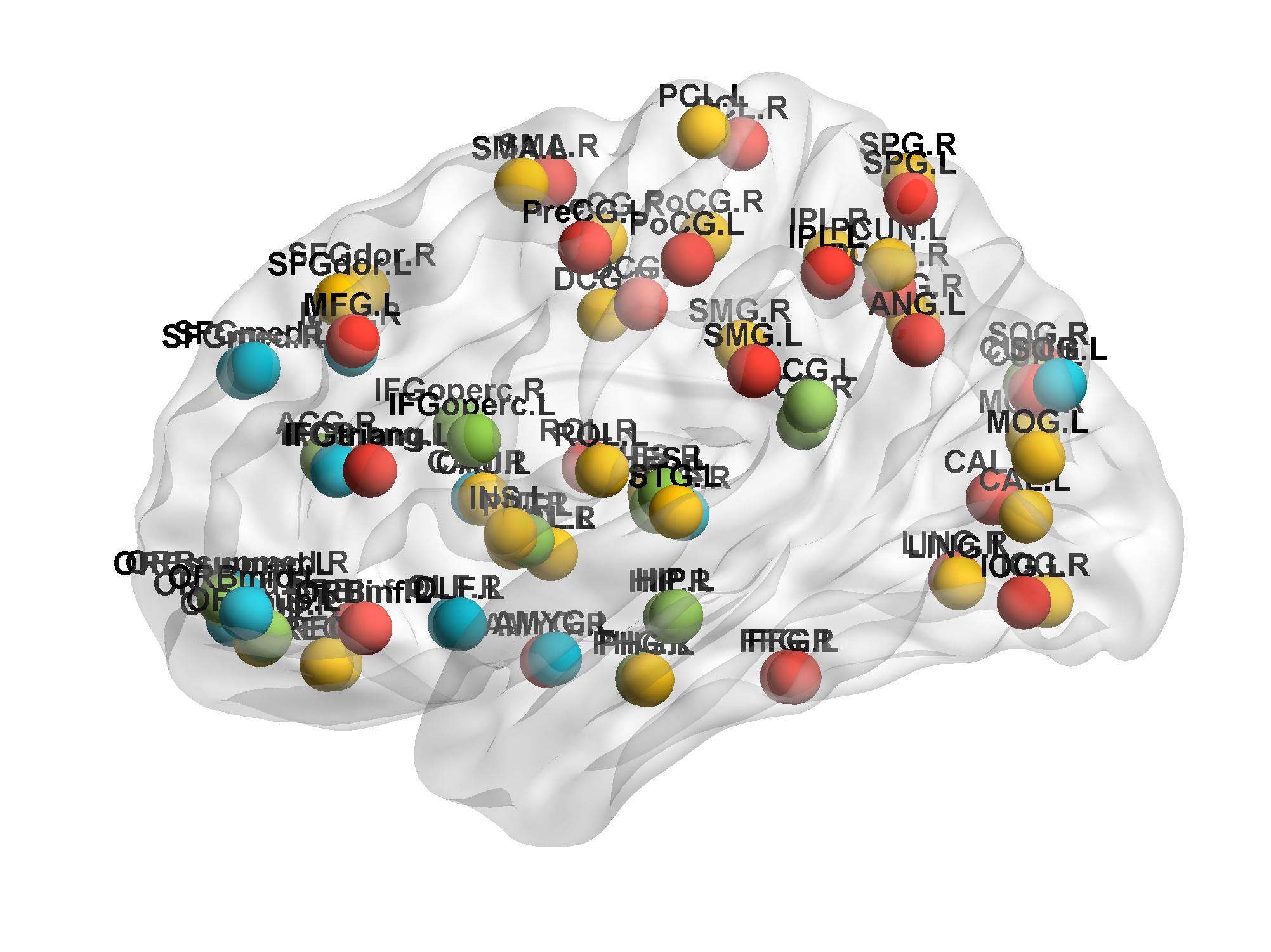}
		    \label{fig:original}
		}%
		\end{subfigure}
		\begin{subfigure}[\scriptsize {Healthy}]{
		\includegraphics[width=.45\linewidth]{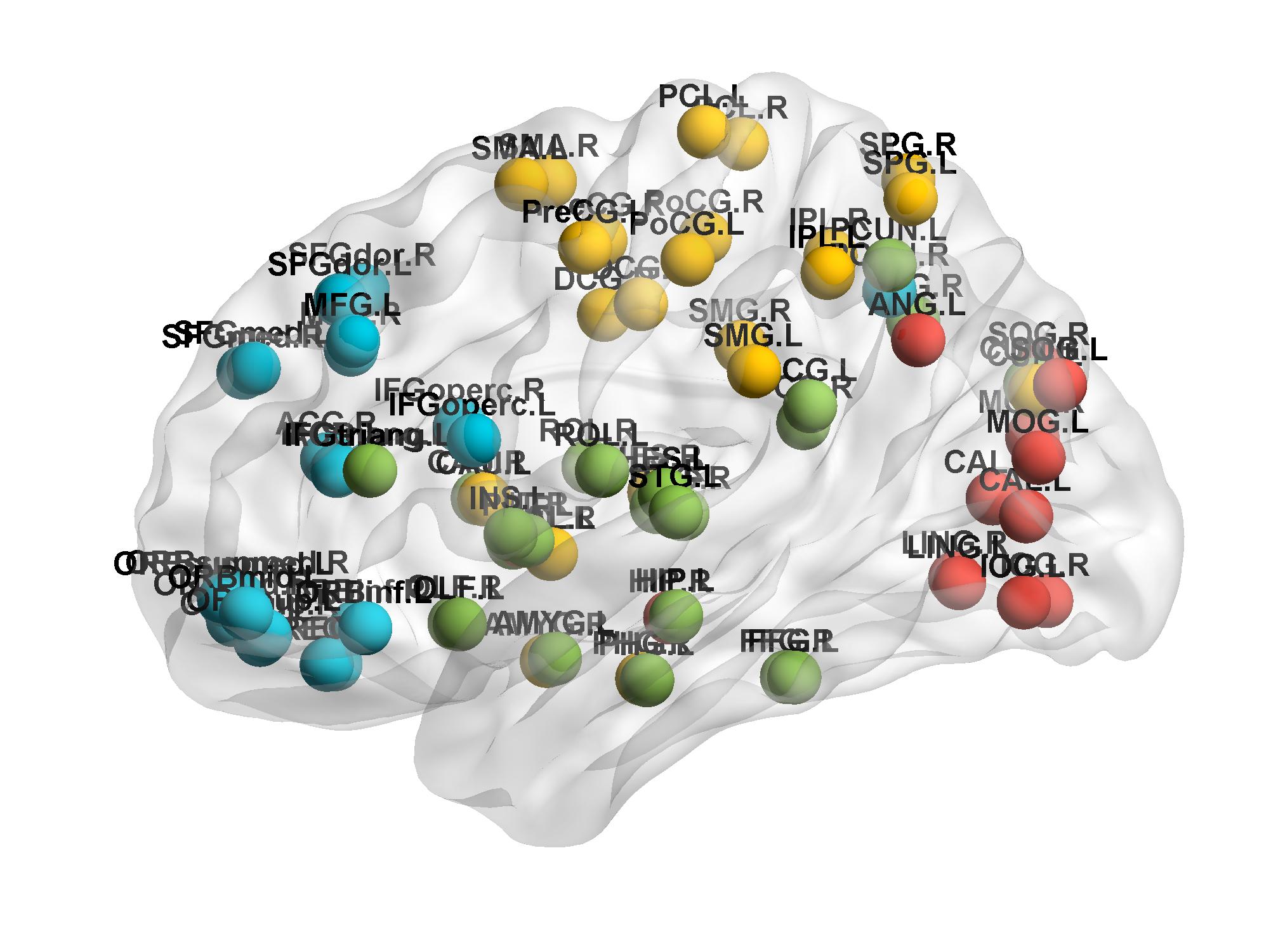}
		\label{fig:afterRW}
		}%
		\end{subfigure}
		\caption{\small{Visualization of the community structure captured by HS-GCN in healthy and bipolar disease networks. Notably this figure highlights the reduced functional connectivity as shown by decreased clustering in the bipolar network.}}\label{fig:brain_network}
\end{figure}

\paragraph{\textbf{Evaluation of the Community Structure in health and disease.}} To evaluate the effectiveness of HS-GCN for capturing the organization and community structure of brain networks, we investigate the resulting brain network embedding by the higher-order GCN. For each brain network, we cluster the brain regions (nodes) using K-means clustering with $k = 4$, and using their embedding feature vectors in K-means. Figure~\ref{fig:brain_network} shows a visualization~\footnote{We used Brainnet Viewer toolbox \cite{xia2013brainnet}} of the community structure of a Bipolar subject versus a healthy control, and nodes are colored by their corresponding community. 
From Fig~\ref{fig:brain_network}, we observe that the nodes of the healthy brain network are well-grouped into the four clusters, while the cluster boundaries in the Bipolar subject's brain network are blurry and the nodes widely spread out over different clusters. We conjecture this observation is due to the reduced functional connectivity as shown by decreased clustering in bipolar subjects versus healthy control (consistent with prior observations in~\cite{teng2014altered,bassett2009human}).

\begin{figure}[t]
\centering
        \begin{subfigure}[\emph{ABIDE}]{		\includegraphics[width=.45\linewidth] {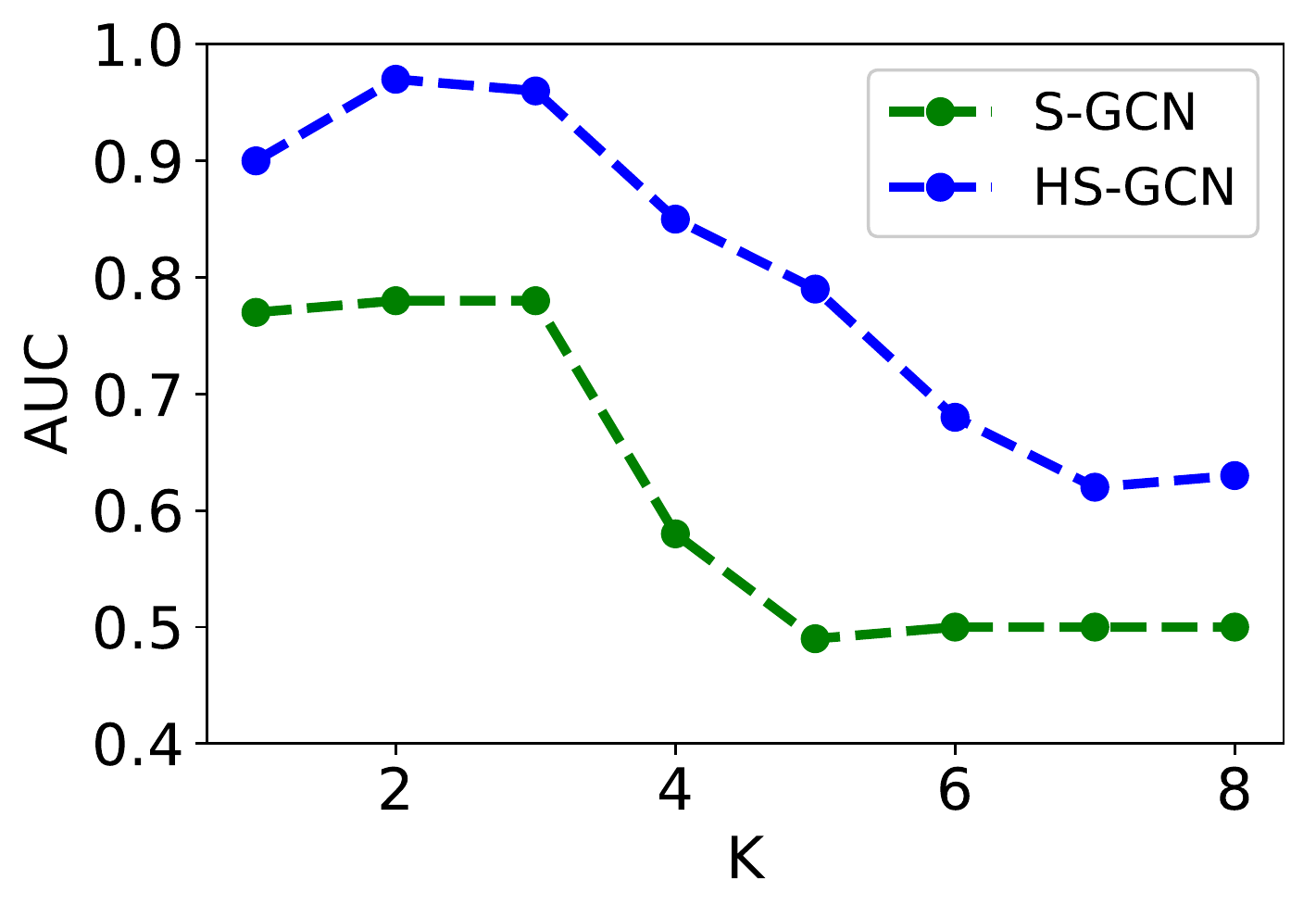}
		    \label{fig:acc_K_ABIDE}
		}%
		\end{subfigure}
	     \hspace{-3mm}
		\begin{subfigure}[\emph{Bipolar}]{
		\includegraphics[width=.45\linewidth]{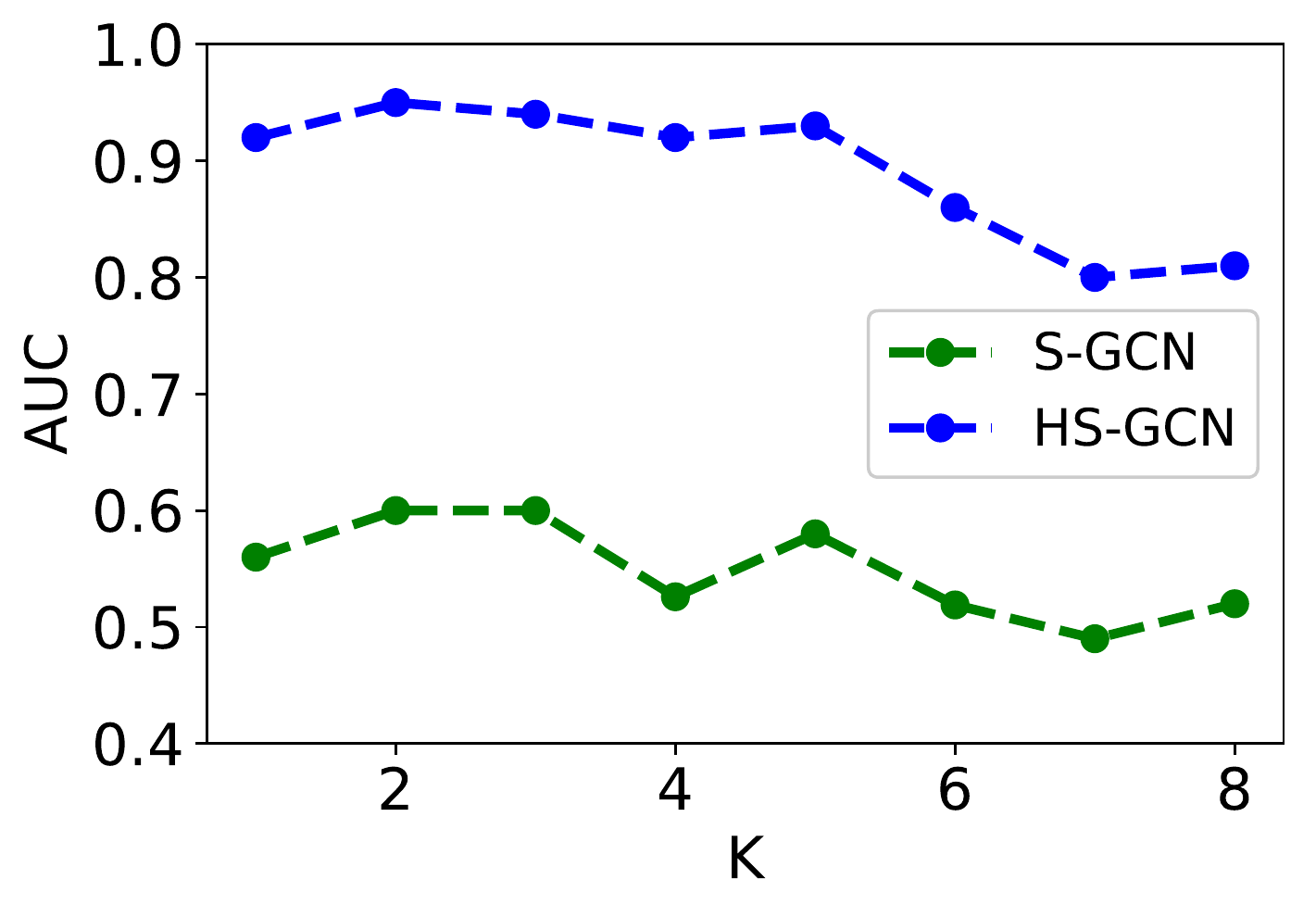}
		\label{fig:acc_K_Bipolar}
		}%
		\end{subfigure}
        \vspace{-4mm}
		\caption{Pair classification AUC of S-GCN and HS-GCN with different values for K}\label{fig:acc_K}
\end{figure}

\begin{figure}[t]
\centering
        \begin{subfigure}[\emph{ABIDE}]{		\includegraphics[width=.45\linewidth] {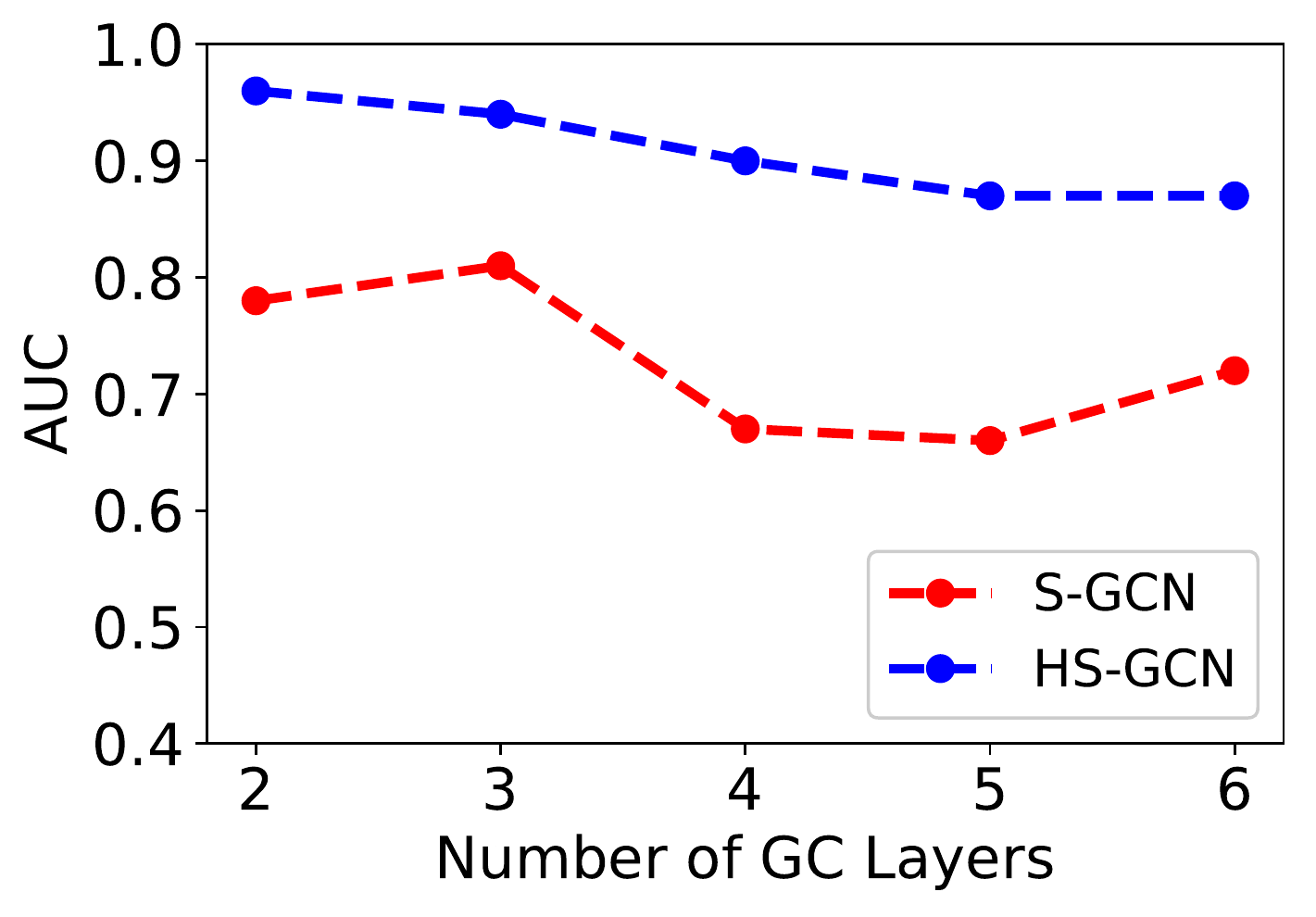}
		    \label{fig:acc_Layers_ABIDE}
		}%
		\end{subfigure}
	   \hspace{-3mm}
		\begin{subfigure}[\emph{Bipolar}]{
		\includegraphics[width=.45\linewidth]{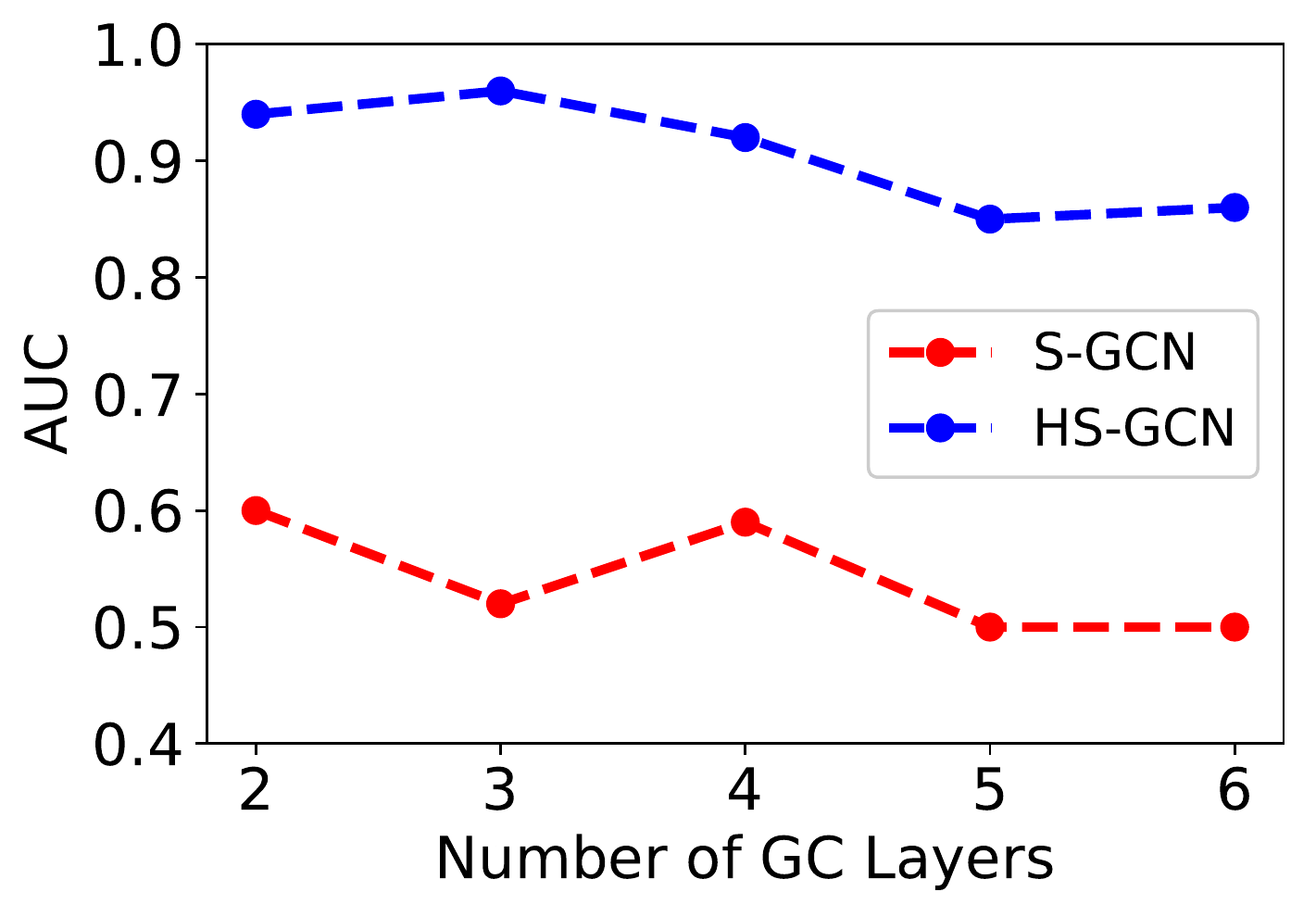}
		\label{fig:acc_Layers_Bipolar}
		}%
		\end{subfigure}
		\vspace{-4mm}
		\caption{Pair classification AUC of S-GCN and HS-GCN with different numbers of GC layers}\label{fig:acc_Layers}
\end{figure}

\paragraph{\textbf{Impact of Higher-order Proximity on GCNs.}} In the GCN models, two important parameters are the number of graph convolutional (GC) layers and the value $K$ which represents the $K$-localized neighborhood considered in the convolutions. In order to evaluate the impact of these parameters on the performance of S-GCN and HS-GCN in pair classification, we apply the two models with various values for $K$ and the number of GC layers. Figure~\ref{fig:acc_K} and Figure~\ref{fig:acc_Layers} show the evaluation results. For the experiments in Figure~\ref{fig:acc_K}, we use two GC layers for both models, and evaluate the models with $K$ values ranging from $1$ to $8$. For the experiments in Figure~\ref{fig:acc_Layers}, we fix $K$ as $3$ and vary the number of GC layers from $2$ to $6$. We set the random walk parameters with the optimal values we found in the experiments above. We can see from Figure~\ref{fig:acc_K} that when the $K$ value increases, the performance of S-GCN first goes up, and when $K$ goes beyond $3$, the AUC scores start to decline. The performance of HS-GCN has a similar trend but it is always better than S-GCN. This indicates that incorporating a larger range of localized neighborhood information in the graph convolutions can not improve the performance of S-GCN as the higher-order property did in HS-GCN. Meanwhile, as we observe in Figure~\ref{fig:acc_Layers}, adding more GC layers did not improve the learning ability of S-GCN either. This is mainly because adding more layers also introduces more hyper-parameters into the model, which could cause the problem of over-fitting. 
\vspace{-2mm}
\paragraph{\textbf{Subject Classification.}}
In this experiment, we use the pairwise similarity learned by the model to further classify the subjects with brain disorder versus healthy controls. We evaluate the proposed HS-GCN model and the baseline S-GCN model on the ABIDE and Bipolar datasets. We apply the weighted $k$-nearest neighbour (kNN) classifier \cite{hechenbichler2004weighted} with the similarity scores we learned for the classification task. The class label of one subject is determined based on a weighted combinations of the labels of its $k$-nearest neighbors. Here we consider all the neighbors with positive similarity scores in the weighted calculation. Meanwhile, we explore the influence of different loss functions in the subject classification performance. Besides evaluating the two models with the Hinge loss in Equation (\ref{hinge_loss}), we also evaluate them with the following constrained variance loss: 
\vspace{-0.5em}
\begin{equation}
\begin{adjustbox}{max width=1\columnwidth}
    $\displaystyle
    \begin{aligned}
         L^{convar}=max(0,{\sigma}^{2+}-a)+max(0,{\sigma}^{2-}-a)+max(0,m-({\mu}^{+}-{\mu}^{-})),
    \end{aligned}
    $
\end{adjustbox}
\label{eq:convar_loss}
\end{equation} where ${\mu}^{+}$ represents the mean similarity between embeddings belonging to the same class, and ${\mu}^{-}$ represents the mean similarity between embeddings belonging to different classes, while ${\sigma}^{2+}$ and ${\sigma}^{2-}$ refer to the variance of pairwise similarity for the same class and different classes, respectively. $m$ is the margin between the means of the same-class and different-class similarity distributions, and $a$ is the variance threshold. This loss function is proposed by \cite{ktena2018metric}. By this formulation, the variance is only penalised when it exceeds the threshold $a$, which allows the similarity estimates to vary around the means, thus could be used to accommodate the diversity that usually exists in fMRI data due to the varied factors in the acquisition process. 
\begin{figure}[t]
\centering
        \begin{subfigure}[\scriptsize{\emph{Hinge Loss}}]{		\includegraphics[width=.44\linewidth] {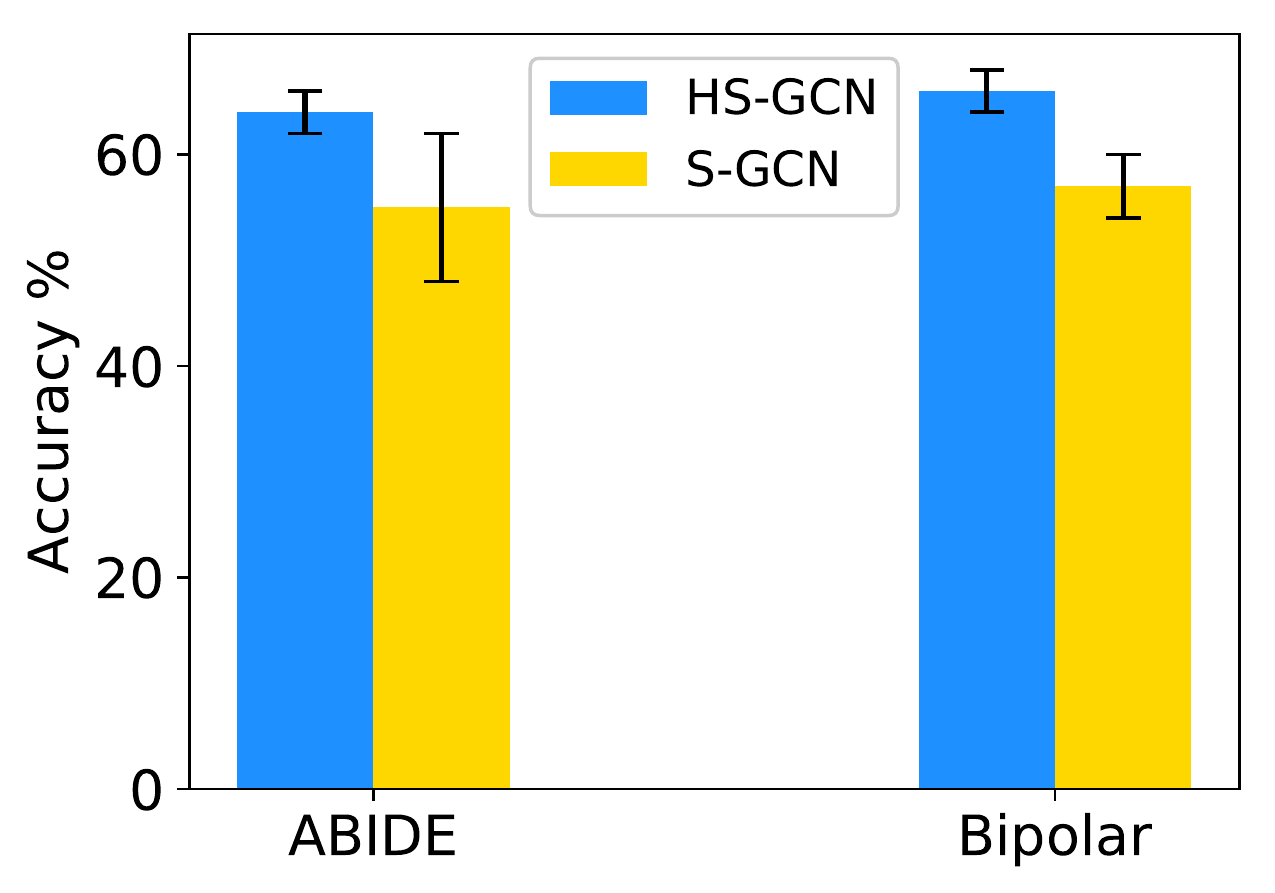}
		    \label{fig:acc_hinge}
		}%
		\end{subfigure}
		\begin{subfigure}[\scriptsize{\emph{Constrained Variance Loss}}]{
		\includegraphics[width=.44\linewidth]{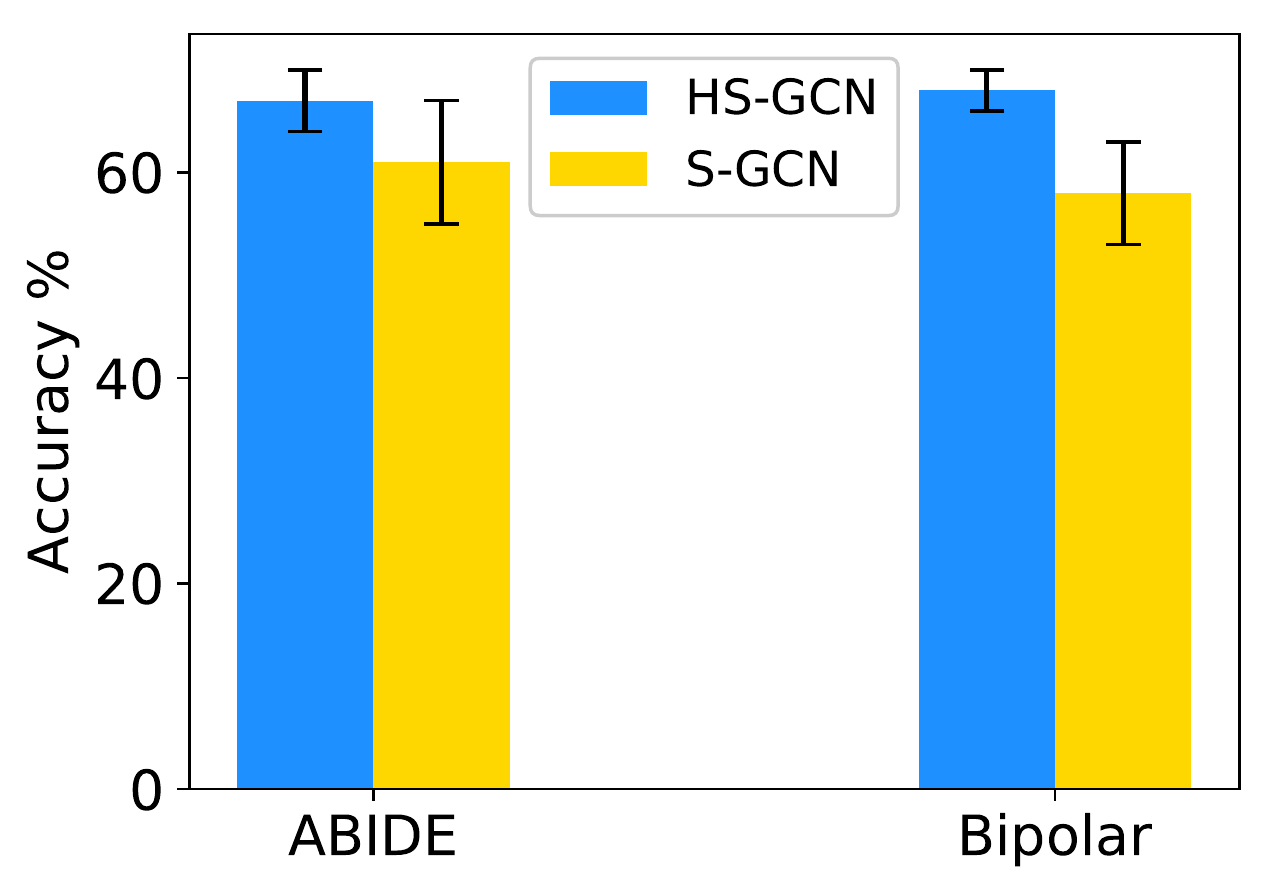}
		\label{fig:acc_convar}
		}%
		\end{subfigure}
		\vspace{-3mm}
		\caption{\small{Subject classification accuracy on ABIDE and Bipolar with two different loss functions}}\label{fig:acc_losses}
		\vspace{-4.mm}
\end{figure}
Figure~\ref{fig:acc_losses} shows the evaluation results of subject classification by the two models with different loss functions. As shown in the Figure, the proposed HS-GCN model achieves a higher accuracy with both loss functions on both datasets compared to the baseline S-GCN model. This implies that the similarity scores learned by the proposed model are more accurate, and thus more reliable to be used for further multi-subject brain connectivity analysis. By comparing Figure~\ref{fig:acc_hinge} and Figure~\ref{fig:acc_convar}, we can find that both models get higher classification accuracy with the similarity scores learned by the constrained variance loss. This could be the benefit from allowing for more diversity across the samples by the constrained variance loss.

\begin{figure}[t]
\centering
        \begin{subfigure}[\scriptsize{\emph{ABIDE}}]{		\includegraphics[width=.45\linewidth] {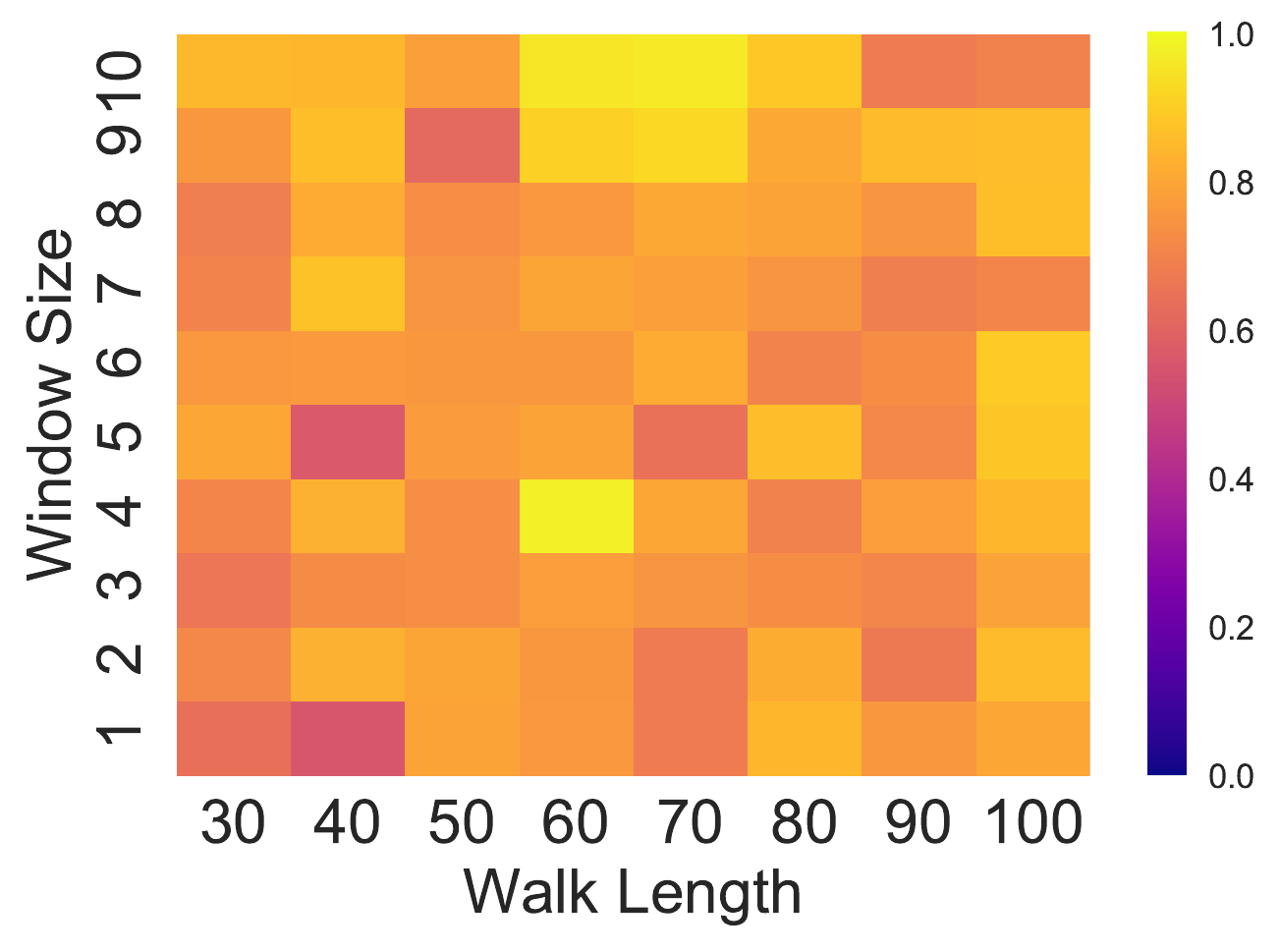}
		    \label{fig:para_ABIDE1}
		}%
		\end{subfigure}
		\begin{subfigure}[\scriptsize{\emph{Bipolar}}]{
		\includegraphics[width=.45\linewidth]{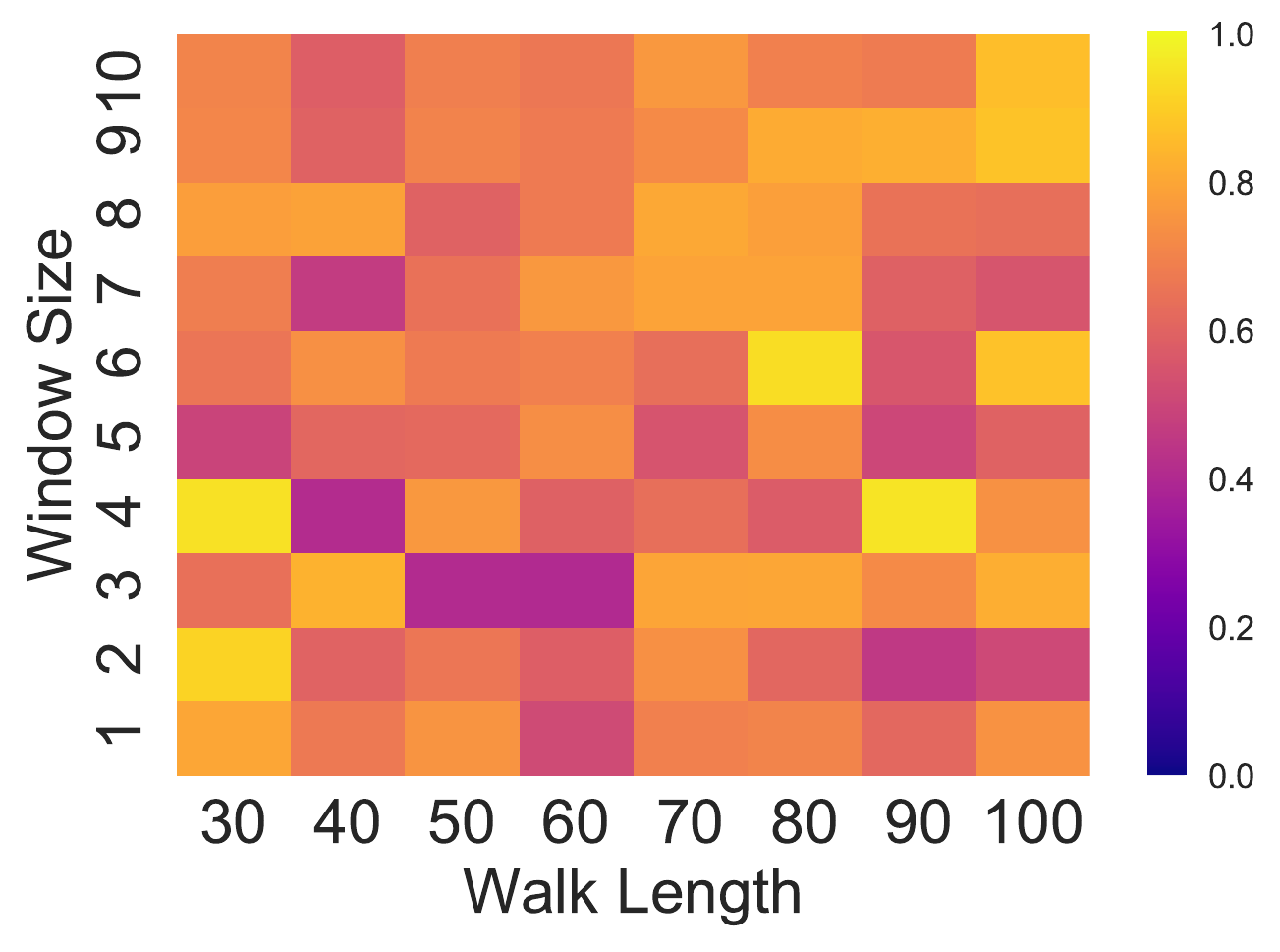}
		\label{fig:para_Bipolar1}
		}%
		\end{subfigure}
		\begin{subfigure}[\scriptsize{\emph{HIV}}]{		
\includegraphics[width=.45\linewidth] {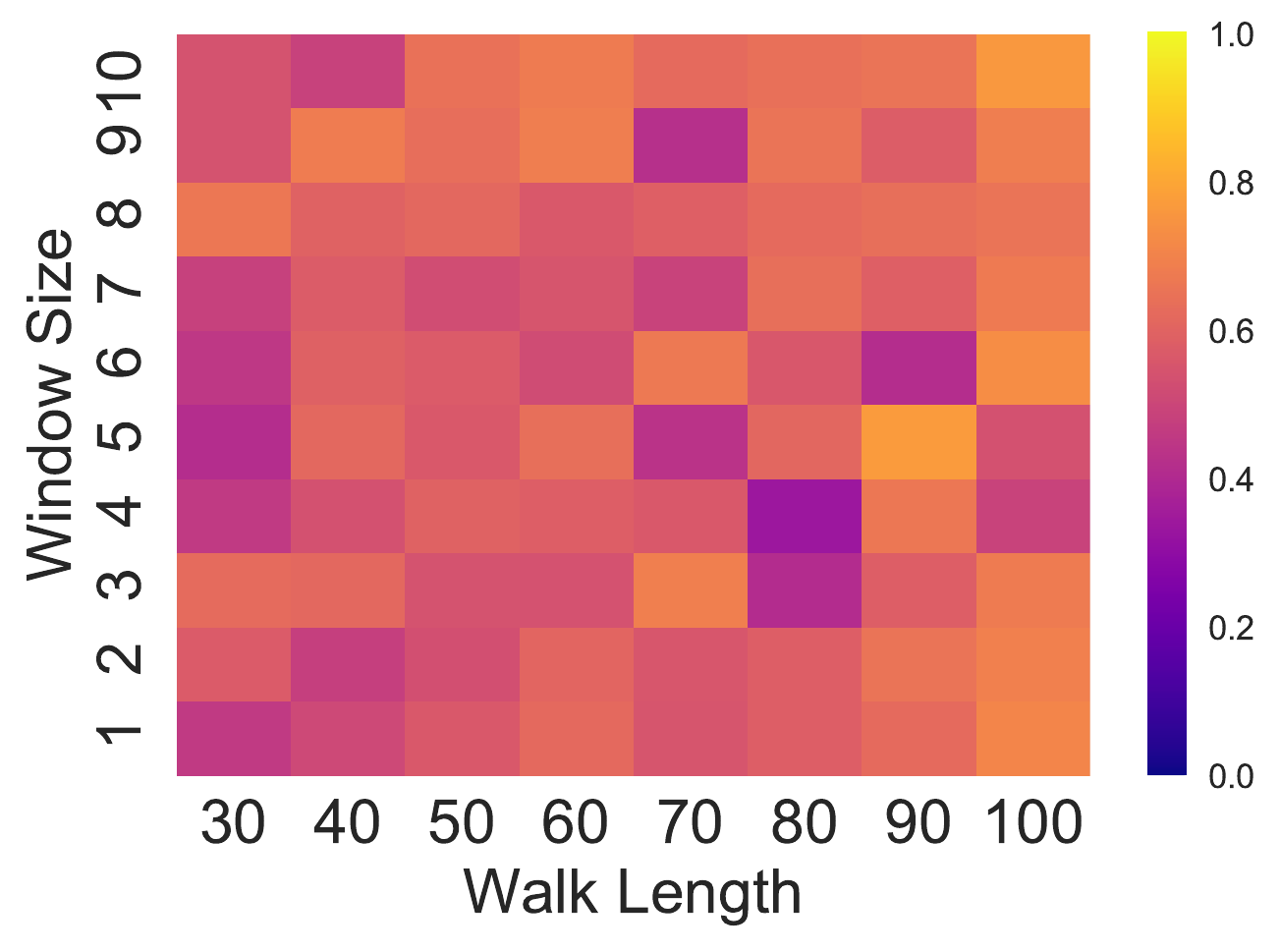}
		    \label{fig:para_HIV1}
		}%
		\end{subfigure}
		\begin{subfigure}[\scriptsize{\emph{HCP}}]{
		\includegraphics[width=.45\linewidth]{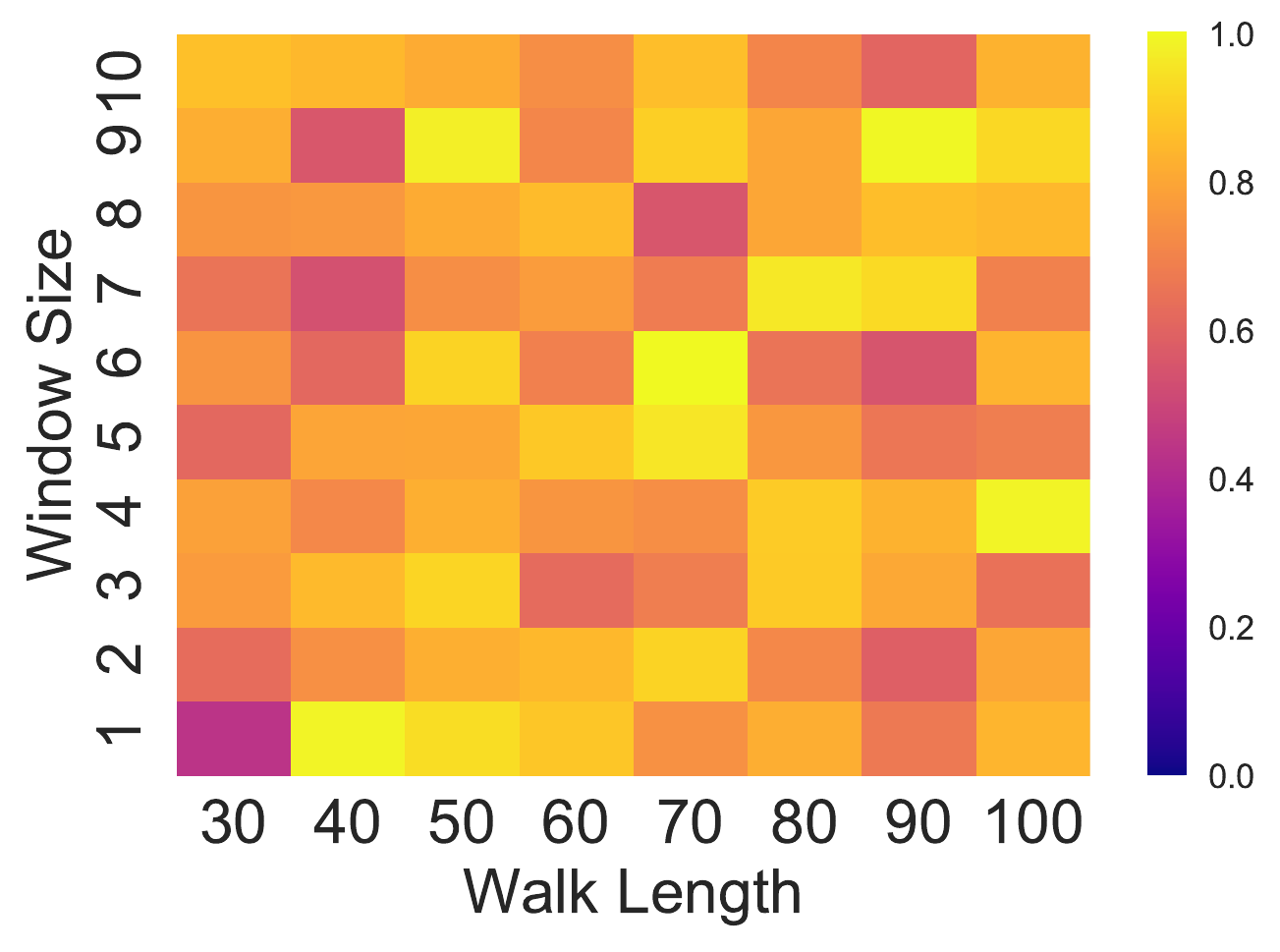}
		\label{fig:para_HCP1}
		}%
		\end{subfigure}
        \vspace{-4mm}
		\caption{\small{AUC scores for Pair classification with different values for Random Walk parameters}}\label{fig:para} 
		\vspace{-4.mm}
\end{figure}

\paragraph{\textbf{Parameter Analysis.}} To analyze the influence of the parameters in the random walk process on the similarity learning performance of HS-GCN, we perform a parameter sensitivity evaluation for the two key parameters in random walk: the walk/path length $l$ and the window size $w$. The bar plots in Figure~\ref{fig:para} show the AUC scores of the proposed model with different values for $l$ and $w$. We observe that the AUC scores vary across different parameter settings. The highest AUC scores on all datasets tend to come from the cases with longer walk length and relatively larger window size. We conjecture this is due to the importance of the long-range functional connectivity for multi-subject brain analysis. For instance, the best result on ABIDE is achieved when $l = 60$ and $w = 4$, and the best result on HIV is achieved when $l = 90$ and $w = 9$. The selection of the parameter values also relates to the scale of the brain networks involved. Among the four datasets, HCP has the largest number of nodes. The optimal walk length value and window size are also the largest among the four datasets. This is reasonable, as the random walks generated on larger graph tend to be longer than those on smaller graphs assuming the two graphs have the same density. To capture the high-order information from larger graphs, we should slide a relatively window on the walks as well. Therefore, it is important to consider the scale and other relevant properties of the brain networks when selecting parameter values.

\section{Related Work}

Our work relates to several branches of studies, which include similarity learning, brain network analysis, and graph convolutional networks. 

Similarity learning is at the heart and foundation of various machine learning tasks, e.g., classification, prediction etc. 
Most of the existing works in this area focus on images~\cite{bromley1994signature,chopra2005learning,koch2015siamese,rossi2018similarity}, while the similarity learning on graph data is seldom studied. Existing methods for similarity estimation between graphs are mainly based on graph embedding, graph kernels or motifs \cite{livi2013graph,rossi2018relational,ijcai2018-456}. These methods are designed for specific scenarios and have their limitations. For example, the graph embedding learned in \cite{abraham2017deriving} may discard structural information that could be important for similarity estimation. In \cite{takerkart2014graph}, the graph kernels used for brain network comparison focus on features of small subgraphs, which ignored global structures of graph. In these works, the graph feature extraction and similarity estimation are done in completely separate stages, where the features extracted may not be suitable for similarity estimation. 
In this paper, we build an end-to-end similarity learning framework to solve these problems.

Brain network analysis has been an emerging research area\cite{liu2017complex}. Existing works in brain networks mainly focus on discovering brain network from spatio-temporal voxel-level data or mining from brain networks for neurological analysis~\cite{bai2017unsupervised,wang2017structural,ma2017multi,safavi2017scalable}. For example, in \cite{bai2017unsupervised}, an unsupervised matrix tri-factorization method is developed to simultaneously discover nodes and edges of the underlying brain networks in fMRI data. \cite{ma2017multib} proposes a multi-view graph embedding approach which learns a unified network embedding from functional and structural brain networks as well as hubs for brain disorder analysis. In \cite{wang2017structural}, a CNN-based model is proposed to learn non-linear structures from brain networks for brain disorder diagnosis. Most of these works aim to learn discriminative features from brain networks for the classification or clustering of subjects. However, how to measure the similarity in the graph domain is seldom studied for multi-subject brain network analysis. In \cite{ktena2018metric}, the metric learning of brain connectivity network is first studied and they use the GCNs to learn features from brain networks. However, they use the traditional GCN model which could not capture the higher-order community structure in brain networks that are important for neurological disorder analysis. In our work, we solve this problem by proposing the higher-order GCNs to characterize the community structure of brain networks, which leads to a much higher performance in similarity learning compared to the method used in \cite{ktena2018metric}.

Graph convolutional network (GCN), first proposed in \cite{defferrard2016convolutional} has now been widely studied. In \cite{kipf2016semi}, a renormalization trick is introduced to simplify computations of GCNs for semi-supervised classification. \cite{li2018deeper} proposes co-training and self-training approaches to improve the training of GCNs in learning with very few labels. In this work, we propose higher-order GCN and employ it in Siamese architecture for similarity learning of brain networks. As the spectral filters in GCNs are closely relies on the graph Laplacian, it is very important to find a good graph representation to be used in the framework. That motivates us to design the higher-order GCNs that operate on higher-order proximity representations of graphs. In future work, we aim to explore the impact of attention networks on both pair and subject classification~\cite{lee2018attention,velivckovic2017graph,lee2018graph,nguyen2018continuous}.  

\section{Conclusion}
We proposed an end-to-end framework called \emph{Higher-order Siamese GCN} (HS-GCN) for learning similarity among fMRI brain networks using higher-order GCNs as the twin networks. The proposed higher-order graph convolutions leverage the higher-order community structure of brain networks. We use this approach to characterize the community and organizational structure in brain networks. To the best of our knowledge, this is the first community-preserving similarity learning framework for multi-subject brain network analysis. Experimental results on four real fMRI datasets demonstrate the potential use cases of the proposed framework for multi-subject brain analysis in health and neuropsychiatric disorders.

\small{

}

\begin{thebibliography}{plain}
\bibliographystyle{aaai}

\bibitem[\protect\citeauthoryear{Abraham \bgroup et al\mbox.\egroup
  }{2017}]{abraham2017deriving}
Abraham, A.; Milham, M.~P.; Di~Martino, A.; Craddock, R.~C.; Samaras, D.;
  Thirion, B.; and Varoquaux, G.
\newblock 2017.
\newblock Deriving reproducible biomarkers from multi-site resting-state data:
  An autism-based example.
\newblock {\em NeuroImage} 147:736--745.

\bibitem[\protect\citeauthoryear{Ahmed \bgroup et al\mbox.\egroup
  }{2015}]{ahmed2015efficient}
Ahmed, N.~K.; Neville, J.; Rossi, R.~A.; and Duffield, N.
\newblock 2015.
\newblock Efficient graphlet counting for large networks.
\newblock In {\em ICDM},  1--10.
\newblock IEEE.

\bibitem[\protect\citeauthoryear{Ahmed \bgroup et al\mbox.\egroup
  }{2018}]{ahmed2018learning}
Ahmed, N.~K.; Rossi, R.; Lee, J.~B.; Kong, X.; Willke, T.~L.; Zhou, R.; and
  Eldardiry, H.
\newblock 2018.
\newblock Learning role-based graph embeddings.
\newblock {\em ICML workshop on StarAI}.

\bibitem[\protect\citeauthoryear{Ahmed, Duffield, and
  Xia}{2018}]{ijcai2018-456}
Ahmed, N.; Duffield, N.; and Xia, L.
\newblock 2018.
\newblock Sampling for approximate bipartite network projection.
\newblock In {\em Proceedings of the Twenty-Seventh International Joint
  Conference on Artificial Intelligence, {IJCAI-18}},  3286--3292.
\newblock International Joint Conferences on Artificial Intelligence
  Organization.

\bibitem[\protect\citeauthoryear{Bai \bgroup et al\mbox.\egroup
  }{2017}]{bai2017unsupervised}
Bai, Z.; Walker, P.; Tschiffely, A.; Wang, F.; and Davidson, I.
\newblock 2017.
\newblock Unsupervised network discovery for brain imaging data.
\newblock In {\em SIGKDD},  55--64.
\newblock ACM.

\bibitem[\protect\citeauthoryear{Bassett and Bullmore}{2009}]{bassett2009human}
Bassett, D.~S., and Bullmore, E.~T.
\newblock 2009.
\newblock Human brain networks in health and disease.
\newblock {\em Current opinion in neurology} 22(4):340.

\bibitem[\protect\citeauthoryear{Bautista, Sanakoyeu, and
  Ommer}{2017}]{bautista2017deep}
Bautista, M.~{\'A}.; Sanakoyeu, A.; and Ommer, B.
\newblock 2017.
\newblock Deep unsupervised similarity learning using partially ordered sets.
\newblock In {\em CVPR},  1923--1932.

\bibitem[\protect\citeauthoryear{Belkin and Niyogi}{2003}]{belkin2003laplacian}
Belkin, M., and Niyogi, P.
\newblock 2003.
\newblock Laplacian eigenmaps for dimensionality reduction and data
  representation.
\newblock {\em Neural computation} 15(6):1373--1396.

\bibitem[\protect\citeauthoryear{Benson, Gleich, and
  Leskovec}{2016}]{benson2016higher}
Benson, A.~R.; Gleich, D.~F.; and Leskovec, J.
\newblock 2016.
\newblock Higher-order organization of complex networks.
\newblock {\em Science} 353(6295):163--166.

\bibitem[\protect\citeauthoryear{Bertolero, Yeo, and
  D’Esposito}{2015}]{bertolero2015modular}
Bertolero, M.~A.; Yeo, B.~T.; and D’Esposito, M.
\newblock 2015.
\newblock The modular and integrative functional architecture of the human
  brain.
\newblock {\em Proceedings of the National Academy of Sciences}
  112(49):E6798--E6807.

\bibitem[\protect\citeauthoryear{Bookheimer~et
  al.}{2018}]{bookheimer2018lifespan}
Bookheimer~et al., S.~Y.
\newblock 2018.
\newblock The lifespan human connectome project in aging: An overview.
\newblock {\em NeuroImage}.

\bibitem[\protect\citeauthoryear{Bromley \bgroup et al\mbox.\egroup
  }{1994}]{bromley1994signature}
Bromley, J.; Guyon, I.; LeCun, Y.; S{\"a}ckinger, E.; and Shah, R.
\newblock 1994.
\newblock Signature verification using a" siamese" time delay neural network.
\newblock In {\em NeurIPS},  737--744.

\bibitem[\protect\citeauthoryear{Bruna \bgroup et al\mbox.\egroup
  }{2013}]{bruna2013spectral}
Bruna, J.; Zaremba, W.; Szlam, A.; and LeCun, Y.
\newblock 2013.
\newblock Spectral networks and locally connected networks on graphs.
\newblock {\em arXiv preprint arXiv:1312.6203}.

\bibitem[\protect\citeauthoryear{Cao \bgroup et al\mbox.\egroup
  }{2015}]{cao2015identification}
Cao, B.; Zhan, L.; Kong, X.; Yu, P.~S.; Vizueta, N.; Altshuler, L.~L.; and
  Leow, A.~D.
\newblock 2015.
\newblock Identification of discriminative subgraph patterns in fmri brain
  networks in bipolar affective disorder.
\newblock In {\em International Conference on Brain Informatics and Health},
  105--114.
\newblock Springer.

\bibitem[\protect\citeauthoryear{Chechik \bgroup et al\mbox.\egroup
  }{2009}]{chechik2009online}
Chechik, G.; Shalit, U.; Sharma, V.; and Bengio, S.
\newblock 2009.
\newblock An online algorithm for large scale image similarity learning.
\newblock In {\em NeurIPS},  306--314.

\bibitem[\protect\citeauthoryear{Chopra, Hadsell, and
  LeCun}{2005}]{chopra2005learning}
Chopra, S.; Hadsell, R.; and LeCun, Y.
\newblock 2005.
\newblock Learning a similarity metric discriminatively, with application to
  face verification.
\newblock In {\em CVPR}, volume~1,  539--546.
\newblock IEEE.

\bibitem[\protect\citeauthoryear{Defferrard, Bresson, and
  Vandergheynst}{2016}]{defferrard2016convolutional}
Defferrard, M.; Bresson, X.; and Vandergheynst, P.
\newblock 2016.
\newblock Convolutional neural networks on graphs with fast localized spectral
  filtering.
\newblock In {\em NeurIPS},  3844--3852.

\bibitem[\protect\citeauthoryear{Di~Martino \bgroup et al\mbox.\egroup
  }{2014}]{di2014autism}
Di~Martino, A.; Yan, C.-G.; Li, Q.; Denio, E.; Castellanos, F.~X.; Alaerts, K.;
  Anderson, J.~S.; Assaf, M.; Bookheimer, S.~Y.; Dapretto, M.; et~al.
\newblock 2014.
\newblock The autism brain imaging data exchange: towards a large-scale
  evaluation of the intrinsic brain architecture in autism.
\newblock {\em Molecular psychiatry} 19(6):659.

\bibitem[\protect\citeauthoryear{Frey and Dueck}{2007}]{frey2007clustering}
Frey, B.~J., and Dueck, D.
\newblock 2007.
\newblock Clustering by passing messages between data points.
\newblock {\em science} 315(5814):972--976.

\bibitem[\protect\citeauthoryear{Greicius}{2008}]{greicius2008resting}
Greicius, M.
\newblock 2008.
\newblock Resting-state functional connectivity in neuropsychiatric disorders.
\newblock {\em Current opinion in neurology} 21(4):424--430.

\bibitem[\protect\citeauthoryear{Guo \bgroup et al\mbox.\egroup
  }{2001}]{guo2001learning}
Guo, G.-D.; Jain, A.~K.; Ma, W.-Y.; and Zhang, H.-J.
\newblock 2001.
\newblock Learning similarity measure for natural image retrieval with
  relevance feedback.
\newblock In {\em CVPR}, volume~1,  I--I.
\newblock IEEE.

\bibitem[\protect\citeauthoryear{Hagmann \bgroup et al\mbox.\egroup
  }{2008}]{hagmann2008mapping}
Hagmann, P.; Cammoun, L.; Gigandet, X.; Meuli, R.; Honey, C.~J.; Wedeen, V.~J.;
  and Sporns, O.
\newblock 2008.
\newblock Mapping the structural core of human cerebral cortex.
\newblock {\em PLoS biology} 6(7):e159.

\bibitem[\protect\citeauthoryear{Hechenbichler and
  Schliep}{2004}]{hechenbichler2004weighted}
Hechenbichler, K., and Schliep, K.
\newblock 2004.
\newblock Weighted k-nearest-neighbor techniques and ordinal classification.

\bibitem[\protect\citeauthoryear{Honey \bgroup et al\mbox.\egroup
  }{2009}]{honey2009predicting}
Honey, C.; Sporns, O.; Cammoun, L.; Gigandet, X.; Thiran, J.-P.; Meuli, R.; and
  Hagmann, P.
\newblock 2009.
\newblock Predicting human resting-state functional connectivity from
  structural connectivity.
\newblock {\em Proceedings of the National Academy of Sciences}
  106(6):2035--2040.

\bibitem[\protect\citeauthoryear{Kingma and Ba}{2014}]{kingma2014adam}
Kingma, D.~P., and Ba, J.
\newblock 2014.
\newblock Adam: A method for stochastic optimization.
\newblock {\em arXiv preprint arXiv:1412.6980}.

\bibitem[\protect\citeauthoryear{Kipf and Welling}{2016}]{kipf2016semi}
Kipf, T.~N., and Welling, M.
\newblock 2016.
\newblock Semi-supervised classification with graph convolutional networks.
\newblock {\em arXiv preprint arXiv:1609.02907}.

\bibitem[\protect\citeauthoryear{Koch, Zemel, and
  Salakhutdinov}{2015}]{koch2015siamese}
Koch, G.; Zemel, R.; and Salakhutdinov, R.
\newblock 2015.
\newblock Siamese neural networks for one-shot image recognition.
\newblock In {\em ICML Deep Learning Workshop}, volume~2.

\bibitem[\protect\citeauthoryear{Ktena \bgroup et al\mbox.\egroup
  }{2018}]{ktena2018metric}
Ktena, S.~I.; Parisot, S.; Ferrante, E.; Rajchl, M.; Lee, M.; Glocker, B.; and
  Rueckert, D.
\newblock 2018.
\newblock Metric learning with spectral graph convolutions on brain
  connectivity networks.
\newblock {\em NeuroImage} 169:431--442.

\bibitem[\protect\citeauthoryear{Lee \bgroup et al\mbox.\egroup
  }{2018}]{lee2018attention}
Lee, J.~B.; Rossi, R.~A.; Kim, S.; Ahmed, N.~K.; and Koh, E.
\newblock 2018.
\newblock Attention models in graphs: A survey.
\newblock {\em arXiv preprint arXiv:1807.07984}.

\bibitem[\protect\citeauthoryear{Lee, Rossi, and Kong}{2018}]{lee2018graph}
Lee, J.~B.; Rossi, R.; and Kong, X.
\newblock 2018.
\newblock Graph classification using structural attention.
\newblock In {\em SIGKDD},  1666--1674.
\newblock ACM.

\bibitem[\protect\citeauthoryear{Li, Han, and Wu}{2018}]{li2018deeper}
Li, Q.; Han, Z.; and Wu, X.-M.
\newblock 2018.
\newblock Deeper insights into graph convolutional networks for semi-supervised
  learning.
\newblock {\em arXiv preprint arXiv:1801.07606}.

\bibitem[\protect\citeauthoryear{Liu \bgroup et al\mbox.\egroup
  }{2017}]{liu2017complex}
Liu, J.; Li, M.; Pan, Y.; Lan, W.; Zheng, R.; Wu, F.-X.; and Wang, J.
\newblock 2017.
\newblock Complex brain network analysis and its applications to brain
  disorders: a survey.
\newblock {\em Complexity} 2017.

\bibitem[\protect\citeauthoryear{Livi and Rizzi}{2013}]{livi2013graph}
Livi, L., and Rizzi, A.
\newblock 2013.
\newblock The graph matching problem.
\newblock {\em Pattern Analysis and Applications} 16(3):253--283.

\bibitem[\protect\citeauthoryear{Ma \bgroup et al\mbox.\egroup
  }{2016}]{ma2016multi}
Ma, G.; He, L.; Cao, B.; Zhang, J.; Philip, S.~Y.; and Ragin, A.~B.
\newblock 2016.
\newblock Multi-graph clustering based on interior-node topology with
  applications to brain networks.
\newblock In {\em ECML PKDD},  476--492.
\newblock Springer.

\bibitem[\protect\citeauthoryear{Ma \bgroup et al\mbox.\egroup
  }{2017a}]{ma2017multi}
Ma, G.; He, L.; Lu, C.-T.; Shao, W.; Yu, P.~S.; Leow, A.~D.; and Ragin, A.~B.
\newblock 2017a.
\newblock Multi-view clustering with graph embedding for connectome analysis.
\newblock In {\em CIKM},  127--136.
\newblock ACM.

\bibitem[\protect\citeauthoryear{Ma \bgroup et al\mbox.\egroup
  }{2017b}]{ma2017multib}
Ma, G.; Lu, C.-T.; He, L.; Philip, S.~Y.; and Ragin, A.~B.
\newblock 2017b.
\newblock Multi-view graph embedding with hub detection for brain network
  analysis.
\newblock In {\em ICDM},  967--972.
\newblock IEEE.

\bibitem[\protect\citeauthoryear{Nguyen \bgroup et al\mbox.\egroup
  }{2018}]{nguyen2018continuous}
Nguyen, G.~H.; Lee, J.~B.; Rossi, R.~A.; Ahmed, N.~K.; Koh, E.; and Kim, S.
\newblock 2018.
\newblock Continuous-time dynamic network embeddings.
\newblock In {\em Companion of the The Web Conference 2018 on The Web
  Conference 2018},  969--976.
\newblock International World Wide Web Conferences Steering Committee.

\bibitem[\protect\citeauthoryear{Perozzi, Al-Rfou, and
  Skiena}{2014}]{perozzi2014deepwalk}
Perozzi, B.; Al-Rfou, R.; and Skiena, S.
\newblock 2014.
\newblock Deepwalk: Online learning of social representations.
\newblock In {\em SIGKDD},  701--710.
\newblock ACM.

\bibitem[\protect\citeauthoryear{Power \bgroup et al\mbox.\egroup
  }{2011}]{power2011functional}
Power, J.~D.; Cohen, A.~L.; Nelson, S.~M.; Wig, G.~S.; Barnes, K.~A.; Church,
  J.~A.; Vogel, A.~C.; Laumann, T.~O.; Miezin, F.~M.; Schlaggar, B.~L.; et~al.
\newblock 2011.
\newblock Functional network organization of the human brain.
\newblock {\em Neuron} 72(4):665--678.

\bibitem[\protect\citeauthoryear{Ragin \bgroup et al\mbox.\egroup
  }{2012}]{ragin2012structural}
Ragin, A.~B.; Du, H.; Ochs, R.; Wu, Y.; Sammet, C.~L.; Shoukry, A.; and
  Epstein, L.~G.
\newblock 2012.
\newblock Structural brain alterations can be detected early in hiv infection.
\newblock {\em Neurology} 79(24):2328--2334.

\bibitem[\protect\citeauthoryear{Rossi, Ahmed, and Koh}{2018}]{rossi2018higher}
Rossi, R.~A.; Ahmed, N.~K.; and Koh, E.
\newblock 2018.
\newblock Higher-order network representation learning.
\newblock In {\em WWW},  3--4.
\newblock International World Wide Web Conferences Steering Committee.

\bibitem[\protect\citeauthoryear{Rossi \bgroup et al\mbox.\egroup
  }{2018a}]{rossi2018similarity}
Rossi, R.~A.; Ahmed, N.~K.; Eldardiry, H.; and Zhou, R.
\newblock 2018a.
\newblock Similarity-based multi-label learning.
\newblock In {\em 2018 International Joint Conference on Neural Networks
  (IJCNN)},  1--8.
\newblock IEEE.

\bibitem[\protect\citeauthoryear{Rossi \bgroup et al\mbox.\egroup
  }{2018b}]{rossi2018relational}
Rossi, R.~A.; Zhou, R.; Ahmed, N.~K.; and Eldardiry, H.
\newblock 2018b.
\newblock Relational similarity machines (rsm): A similarity-based learning
  framework for graphs.
\newblock In {\em 2018 IEEE International Conference on Big Data (Big Data)},
  1807--1816.
\newblock IEEE.

\bibitem[\protect\citeauthoryear{Rossi, Zhou, and Ahmed}{2017}]{rossi2017deep}
Rossi, R.~A.; Zhou, R.; and Ahmed, N.~K.
\newblock 2017.
\newblock Deep feature learning for graphs.
\newblock {\em arXiv preprint arXiv:1704.08829}.

\bibitem[\protect\citeauthoryear{Safavi, Sripada, and
  Koutra}{2017}]{safavi2017scalable}
Safavi, T.; Sripada, C.; and Koutra, D.
\newblock 2017.
\newblock Scalable hashing-based network discovery.
\newblock In {\em ICDM},  405--414.
\newblock IEEE.

\bibitem[\protect\citeauthoryear{Shuman \bgroup et al\mbox.\egroup
  }{2013}]{shuman2013emerging}
Shuman, D.~I.; Narang, S.~K.; Frossard, P.; Ortega, A.; and Vandergheynst, P.
\newblock 2013.
\newblock The emerging field of signal processing on graphs: Extending
  high-dimensional data analysis to networks and other irregular domains.
\newblock {\em IEEE Signal Processing Magazine} 30(3):83--98.

\bibitem[\protect\citeauthoryear{Smith}{2002}]{smith2002tutorial}
Smith, L.~I.
\newblock 2002.
\newblock A tutorial on principal components analysis.
\newblock Technical report.

\bibitem[\protect\citeauthoryear{Spronk \bgroup et al\mbox.\egroup
  }{2018}]{spronk2018mapping}
Spronk, M.; Ji, J.~L.; Kulkarni, K.; Repovs, G.; Anticevic, A.; and Cole, M.~W.
\newblock 2018.
\newblock Mapping the human brain's cortical-subcortical functional network
  organization.
\newblock {\em bioRxiv}  206292.

\bibitem[\protect\citeauthoryear{Takerkart \bgroup et al\mbox.\egroup
  }{2014}]{takerkart2014graph}
Takerkart, S.; Auzias, G.; Thirion, B.; and Ralaivola, L.
\newblock 2014.
\newblock Graph-based inter-subject pattern analysis of fmri data.
\newblock {\em PloS one} 9(8):e104586.

\bibitem[\protect\citeauthoryear{Teng \bgroup et al\mbox.\egroup
  }{2014}]{teng2014altered}
Teng, S.; Lu, C.-F.; Wang, P.-S.; Li, C.-T.; Tu, P.-C.; Hung, C.-I.; Su, T.-P.;
  and Wu, Y.-T.
\newblock 2014.
\newblock Altered resting-state functional connectivity of striatal-thalamic
  circuit in bipolar disorder.
\newblock {\em PloS one} 9(5):e96422.

\bibitem[\protect\citeauthoryear{Van~Essen \bgroup et al\mbox.\egroup
  }{2013}]{van2013wu}
Van~Essen, D.~C.; Smith, S.~M.; Barch, D.~M.; Behrens, T.~E.; Yacoub, E.;
  Ugurbil, K.; Consortium, W.-M.~H.; et~al.
\newblock 2013.
\newblock The wu-minn human connectome project: an overview.
\newblock {\em Neuroimage} 80:62--79.

\bibitem[\protect\citeauthoryear{Veli{\v{c}}kovi{\'c} \bgroup et
  al\mbox.\egroup }{2017}]{velivckovic2017graph}
Veli{\v{c}}kovi{\'c}, P.; Cucurull, G.; Casanova, A.; Romero, A.; Li{\`o}, P.;
  and Bengio, Y.
\newblock 2017.
\newblock Graph attention networks.
\newblock {\em arXiv preprint arXiv:1710.10903}.

\bibitem[\protect\citeauthoryear{Wang \bgroup et al\mbox.\egroup
  }{2014}]{wang2014learning}
Wang, J.; Song, Y.; Leung, T.; Rosenberg, C.; Wang, J.; Philbin, J.; Chen, B.;
  and Wu, Y.
\newblock 2014.
\newblock Learning fine-grained image similarity with deep ranking.
\newblock In {\em CVPR},  1386--1393.

\bibitem[\protect\citeauthoryear{Wang \bgroup et al\mbox.\egroup
  }{2017}]{wang2017structural}
Wang, S.; He, L.; Cao, B.; Lu, C.-T.; Yu, P.~S.; and Ragin, A.~B.
\newblock 2017.
\newblock Structural deep brain network mining.
\newblock In {\em SIGKDD},  475--484.
\newblock ACM.

\bibitem[\protect\citeauthoryear{Whitfield-Gabrieli and
  Nieto-Castanon}{2012}]{whitfield2012conn}
Whitfield-Gabrieli, S., and Nieto-Castanon, A.
\newblock 2012.
\newblock Conn: a functional connectivity toolbox for correlated and
  anticorrelated brain networks.
\newblock {\em Brain connectivity} 2(3):125--141.

\bibitem[\protect\citeauthoryear{Xia, Wang, and He}{2013}]{xia2013brainnet}
Xia, M.; Wang, J.; and He, Y.
\newblock 2013.
\newblock Brainnet viewer: a network visualization tool for human brain
  connectomics.
\newblock {\em PloS one} 8(7):e68910.


\end{thebibliography}
\end{document}